\documentclass{article}

\usepackage[final]{neurips_2023}

\usepackage{wrapfig}

\usepackage[utf8]{inputenc} % allow utf-8 input
\usepackage[T1]{fontenc}    % use 8-bit T1 fonts
\usepackage{hyperref}       % hyperlinks
\usepackage{url}            % simple URL typesetting
\usepackage{booktabs}       % professional-quality tables
\usepackage{amsfonts}       % blackboard math symbols
\usepackage{nicefrac}       % compact symbols for 1/2, etc.
\usepackage{microtype}      % microtypography
\usepackage{xcolor}         % colors

\usepackage{subcaption}

\usepackage{amsmath}
\usepackage{amssymb}
\usepackage{mathtools}
\usepackage{amsthm}
\usepackage{bm}
\usepackage{adjustbox}

\usepackage[capitalize,noabbrev]{cleveref}

\usepackage{tikz,tikz-cd,forest}
\usetikzlibrary{fit,arrows,calc,positioning,shadows.blur,backgrounds,decorations.markings}

\usepackage{longtable}
\usepackage{multirow}
\usepackage{xspace}
\theoremstyle{plain}

\theoremstyle{definition}

\theoremstyle{remark}

\def\1{\bm{1}}

\newcommand*{\fullref}[1]{\hyperref[{#1}]{\cref{#1} (\nameref*{#1})}}

\newcommand{\eat}[1]{}

\mathchardef\mhyphen="2D % Define a "math hyphen"
\mathchardef\mdash="2D % Define a "math hyphen"

\newcommand{\myvec}[1]{\mathbf{#1}}
\newcommand{\myvecsym}[1]{\boldsymbol{#1}}
\makeatletter
\newcommand{\oset}[3][-0.3ex]{%
  \mathrel{\mathop{#3}\limits^{
    \vbox to#1{\kern-4\ex@
    \hbox{$\scriptstyle#2$}\vss}}}}
\makeatother

\newcommand{\vone}{\myvecsym{1}}

\newcommand{\valpha}{\myvecsym{\alpha}}

\newcommand{\vpi}{\myvecsym{\pi}}

\newcommand{\vtheta}{\myvecsym{\theta}}

\newcommand{\vx}{\myvec{x}}
\newcommand{\vM}{\myvec{M}}

\newcommand{\vX}{\myvec{X}}
\DeclareMathAlphabet{\mathsfit}{\encodingdefault}{\sfdefault}{m}{sl}
\SetMathAlphabet{\mathsfit}{bold}{\encodingdefault}{\sfdefault}{bx}{n}

\newcommand{\mymathcal}[1]{\mathcal{#1}}

\newcommand{\calD}{{\mymathcal{D}}}

\newcommand{\calL}{\mymathcal{L}}

\newcommand{\Dir}{\mathrm{Dir}}

\newcommand{\indep}[2]{{#1} \perp {#2}}

\newcommand{\ra}{\rightarrow}

\newcommand{\const}{\mathrm{const}}

\newcommand{\ind}[1]{\mathbb{I}\left({#1}\right)}

\newcommand{\loss}{\calL}

\newcommand{\data}{\calD}

\newcommand{\be}{\begin{equation}}
\newcommand{\ee}{\end{equation}}
\newcommand{\bea}{\begin{eqnarray}}
\newcommand{\eea}{\end{eqnarray}}
\newcommand{\beaa}{\begin{eqnarray*}}
\newcommand{\eeaa}{\end{eqnarray*}}
\newcommand{\ba}{\begin{align*}}
\newcommand{\ea}{\end{align*}}

\newcommand{\figfix}[1]{}
\newcommand{\addfig}[4] %{options}{caption}{label}{filename}
{
    \begin{figure}
    \centering
    \includegraphics[#1]{\figdir/#4}
    \caption{#2}
    \label{fig:#3}
    \end{figure}
}

\newcommand{\addtwofigs}[5]    %{options}{caption}{label}{fname1}{fname2}
{
    \begin{figure}
    \centering
    \begin{subfigure}{0.45\textwidth}
      \centering
      \includegraphics[#1]{\figdir/#4}
      \caption{ }
     \end{subfigure}
~
    \begin{subfigure}{0.45\textwidth}
            \centering
            \includegraphics[#1]{\figdir/#5}
                  \caption{ }
    \end{subfigure}
    \caption{#2}
    \label{fig:#3}
    \end{figure}
}

\newcommand{\addthreefigs}[6]    %{options}{caption}{label}{fname1}{fname2}{fname3}
{
    \begin{figure}
    \centering
    \begin{subfigure}{0.3\textwidth}
      \centering
      \includegraphics[#1]{\figdir/#4}
      \caption{ }
     \end{subfigure}
~
    \begin{subfigure}{0.3\textwidth}
            \centering
            \includegraphics[#1]{\figdir/#5}
                  \caption{ }
  \end{subfigure}
~
    \begin{subfigure}{0.3\textwidth}
            \centering
            \includegraphics[#1]{\figdir/#6}
                  \caption{ }
    \end{subfigure}
    \caption{#2}
    \label{fig:#3}
    \end{figure}
}

\newcommand{\addfourfigs}[7]    %{options}{caption}{label}{fname1}{fname2}{fname3}{fname4}
{
    \begin{figure}
    \centering
    \begin{subfigure}{0.45\textwidth}
      \centering
      \includegraphics[#1]{\figdir/#4}
      \caption{ }
     \end{subfigure}
~
    \begin{subfigure}{0.45\textwidth}
      \centering
      \includegraphics[#1]{\figdir/#5}
      \caption{ }
     \end{subfigure}
\\
    \begin{subfigure}{0.45\textwidth}
      \centering
      \includegraphics[#1]{\figdir/#6}
      \caption{ }
     \end{subfigure}
~
    \begin{subfigure}{0.45\textwidth}
      \centering
      \includegraphics[#1]{\figdir/#7}
      \caption{ }
     \end{subfigure}
    \caption{#2}
    \label{fig:#3}
    \end{figure}
}

\newcommand{\twodigits}[1]{\ifnum #1 < 10 0\fi #1}

\newcommand{\points}[1]{}

\def\vtheta{{\bm{\theta}}}

\def\vx{{\bm{x}}}

\makeatletter
\newdimen\nodeSize
\newdimen\nodeDist
\tikzset{
    position/.style args={#1:#2 from #3}{
        at=(#3.#1), anchor=#1+180, shift=(#1:#2)
    }
}
\makeatletter

\makeatletter
\let\tikzcd@ar@new@orig\tikzcd@ar@new
\def\tikzcd@ar@new@bg[#1]{% https://tex.stackexchange.com/a/664328/16595
  \pgfutil@g@addto@macro\tikzcd@savedpaths{%
    \pgfonlayer{background}\path[/tikz/commutative diagrams/.cd,every arrow,#1]%
    (\tikzcd@ar@start\tikzcd@startanchor)to(\tikzcd@ar@target\tikzcd@endanchor);
    \endpgfonlayer}\let\tikzcd@ar@new\tikzcd@ar@new@orig}
\tikzcdset{every diagram/.append code=%
  \def\bgarrow{\let\tikzcd@ar@new\tikzcd@ar@new@bg\tikzcd@arrow}%
  }
\makeatother

\tikzset{
  vert down/.style={to path={--(\tikztostart|-\tikztotarget.north)\tikztonodes}},
  lab/.style={label={[anchor=south east]south east:$#1$}},
  lab'/.style={label={[anchor=south west,name=placeholder,xshift=-\pgfkeysvalueof{/pgf/inner xsep}]
    south east:\phantom{$#1$}}},
  submatrix/.style args={[#1]#2:#3}{
    fit={#2},name={#3},node contents=,draw,#1}}
\tikzset{
  roundnode/.style={circle, draw=blue!60, fill=blue!5, very thick},
  squarednode/.style={rectangle, draw=red!60, fill=red!5, very thick}}

\tikzcdset{
  submatrix/.style args={[#1]#2:#3}{
    to path={node[inner sep=+.85ex,name prefix=\tikzcdmatrixname-,submatrix={[{#1}]{#2}:{#3}}]}},
  rows'/.style 2 args={
    /utils/temp/.style={/tikz/row ##1/.append style={nodes={#2}}},/utils/temp/.list={#1}}}

\tikzset{forest edge/.style={style/.expanded=\forestoption{edge}}}

\newcommand{\figdir}{figures}

\newcommand{\psource}{p_s}
\newcommand{\ptarget}{p_t}
\newcommand{\pbalanced}{p_b}
\newcommand{\chexpert}{CheXpert\xspace}
\newcommand{\cmnist}{Colored MNIST\xspace}

\tikzset{lightning bolt to/.style={to path={
let \p1=(\tikztostart), \p2=(\tikztotarget), \n1={veclen(\y2-\y1,\x2-\x1)} in
  (\p1) -- ($($(\p1)!0.6!(\p2)$)!\n1*.1!-90:(\p2)$) -- ($(\p1)!0.55!(\p2)$) --
  (\p2) -- ($($(\p1)!0.4!(\p2)$)!\n1*.1!90:(\p2)$) -- ($(\p1)!0.45!(\p2)$) -- 
  cycle (\p2)% Move to end point
}}}

\title{Beyond Invariance: Test-Time Label-Shift Adaptation for Addressing ``Spurious'' Correlations}

\author{%
  Qingyao Sun\thanks{Work done as a master's student at the University of Chicago.} \\
  Cornell University \\
  \texttt{qs234@cornell.edu} \\
  \And
  Kevin Murphy \\
  Google DeepMind \\
  \texttt{kpmurphy@google.com} \\
  \AND
  Sayna Ebrahimi \\
  Google Cloud AI Research \\
  \texttt{saynae@google.com} \\
  \And
  Alexander D'Amour \\
  Google DeepMind \\
  \texttt{alexdamour@google.com} \\
}

\begin{document}

\maketitle

\begin{abstract}
Changes in the data distribution at test time can have deleterious effects on the performance of predictive models $p(y|x)$.
We consider situations where there are additional meta-data labels (such as group labels), denoted by $z$, that can account for such changes in the distribution. In particular, we assume that 
the prior distribution $p(y,z)$, which models the dependence between the class label $y$ and the ``nuisance'' factors $z$, may change across domains, either due to a change in the correlation between these terms, or a change in one of their marginals. However, we assume that the generative model for features $p(x|y,z)$ is invariant across domains. We note that this corresponds to an expanded version of the widely used
``label shift'' assumption, where the labels now also include the nuisance factors $z$. 
Based on this observation,  we
 propose a test-time label shift correction that adapts to changes in the joint distribution $p(y, z)$ using EM applied to unlabeled samples from the target domain distribution, $\ptarget(x)$.
 Importantly, we are able to avoid fitting a generative model $p(x|y,z)$,
 and merely need to reweight the outputs of a discriminative
 model $\psource(y,z|x)$ trained on the source distribution.
  We evaluate our method, which we call ``Test-Time Label-Shift Adaptation'' (TTLSA), on several standard image and text datasets, as well as the \chexpert chest X-ray dataset, 
and show that it improves performance over methods that target invariance to changes in the distribution,
as well as baseline
empirical risk minimization methods.
Code for reproducing experiments is available at \url{https://github.com/nalzok/test-time-label-shift}.
\end{abstract}
\section{Introduction}

Machine learning systems are known to be sensitive to so-called ``spurious correlations''~\citep{geirhos2020shortcut,wiles2022a,arjovsky2019invariant} between irrelevant features of the inputs and the predicted output label.
These features and their associated correlations are called ``spurious'' because they are expected to change across real-world distribution shifts.
As a result, models that rely on such spurious correlations often have worse performance when they are deployed on a \textit{target domain} that is distinct from the \textit{source domain} on which they were trained~\citep{geirhos2020shortcut,Izmailov2022}.
For example, \citet{Jabbour2020} show that neural networks trained to recognize conditions like pneumonia from chest X-rays can learn to rely on features that are predictive of patient demographics rather than the medical condition itself.
When the correlation between demographic factors and the condition change (e.g., when the model is deployed on a different patient population), the performance of such models suffers.

To address this issue, recent work has focused on learning predictors that are invariant to changes in spurious correlations across source and target domains, either using data from multiple environments \citep[e.g.,][]{arjovsky2019invariant}, or data where 
the labels for the nuisance factors are available at training time \citep{Veitch2021,Makar2022,Puli2022,Makino2022}.

However, ``spurious'' correlations can sometimes provide valuable prior information for examples where the input is ambiguous.
%does not provide conclusive evidence about the target label.
Consider the example of calculating the probability that a patient has a particular disease given a positive test.
It is well known that the underlying prevalence of the disease (i.e., prior probability it is present) in the patient population is highly informative for making this diagnosis; even if the test is highly sensitive and specific, the patient's probability of having the disease given a positive test may be low if the disease is rare given that patient's background \citep[see, e.g.,][]{bours2021bayes}.
A similar logic applies in many prediction problems: if we can predict, say, patient demographics from the input, and the prevalence of the target label differs between demographic groups in the current patient population, then the demographic-relevant spurious features provide prior information that supplements the information in the target-relevant features of the input.
%Thus, when it comes to target domain performance, the problem is not that spurious features are being used; rather, it is that spurious features are being used {\em inappropriately} for that domain.
Thus, if a model could be adjusted to use spurious features optimally in downstream target domains, it could substantially out-perform invariant predictors, as we show in this paper.\footnote{
We acknowledge that practitioners may have concerns beyond target domain performance, such as certain notions of equal treatment, 
in which the reliance on such spurious features is problematic or even forbidden.
%This is orthogonal to the concerns of this paper.
%However, equal treatment criteria are often motivated by a desire to ensure equitable --- and maximal --- downstream performance across groups, no matter what the group demographics, and our method achieves this goal.
However, equal treatment criteria are often motivated by a desire to ensure that models perform as well as possible when they are applied in domains with very different population compositions, 
and our method can achieve this goal \citep[see][for related discussion]{Makar2022fairness}.
}

%\todoqingyao{In case the ``increased fairness'' argument is not convincing enough, we can also enforce fairness (in the strictest sense) by setting the target label prior to a uniform distribution, or doing logit adjustment, or both.}
%\todoalex{This is approximately true, but the issues are sort of subtle. I don't think we should lean too hard into a claim of ``solving'' fairness here. The goal of the footnote is acknowledge that one might want invariance for other \emph{a priori} reasons, so we're not saying that invariance is never good.}

\begin{wrapfigure}{r}{0.38\textwidth}
%\begin{figure}
\centering
\begin{tikzpicture}[
roundnode/.style={circle, draw=blue!60, fill=blue!5, very thick, minimum size=6mm},
invisible/.style={circle, draw=white, fill=white}
]
\node[roundnode] (U) {$U$};
\node[roundnode] (Y) [below left=2mm of U, position=-135:{\nodeDist} from U] {$Y$};
\node[roundnode] (Z) [below right=2mm of U, position=-45:{\nodeDist} from U] {$Z$};
\node[roundnode] (X) [below=14mm of U] {$X$};
\node[invisible] (L) [above right=1mm of U, position=45:{\nodeDist} from U]{};

\draw[->, ultra thick] (U.south west) -- (Y.north east);
\draw[->, ultra thick] (U.south east) -- (Z.north west);
\draw[->, ultra thick] (Y.south east) -- (X.north west);
\draw[->, ultra thick] (Z.south west) -- (X.north east);

\fill[yellow] (L.west) to [lightning bolt to] (U.north);
\draw[black,thick] (L.west) to [lightning bolt to] (U.north);
\end{tikzpicture}
\caption{
  Modeling assumptions.
  $U$ is a hidden confounder
  that induces a spurious correlation between the label
  $Y$ and other causal factors $Z$,
  which together generate the features $X$.
  The distribution of $U$ can change between target and source domains, but the distribution of $X$ given $Y, Z$ is fixed.
}
\label{fig:model}
%\end{figure}
\end{wrapfigure}
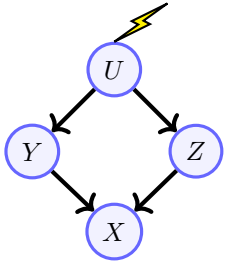

\eat{
%\documentclass[tikz,border=10pt]{standalone}
%\usetikzlibrary{shapes,arrows,positioning}
\usetikzlibrary{shapes,arrows,positioning,decorations.pathmorphing}

\begin{figure}
\begin{tikzpicture}[
    node distance=2cm,
    auto,
    thick,
    main node/.style={circle,draw,font=\sffamily\Large\bfseries}]

    \node[main node] (1) {U};
    \node[main node] (2) [below left = of 1] {Z};
    \node[main node] (3) [below right = of 1] {Y};
    \node[main node] (4) [below right = of 2] {X};

    \path[->]
        (1) edge node {} (2)
        (1) edge node {} (3)
        (2) edge node {} (4)
        (3) edge node {} (4);

 % Add lightning bolt above U
 %   \draw[white,double=black,decoration={zigzag,segment length=3mm,amplitude=.5mm},decorate,very thick] ($(1)+(0,.5)$) -- ++(0,.5);
% This code now includes a lightning bolt symbol on top of the U node. Please note that the $(1)+(0,.5)$ refers to a point which is .5 units above the center of the node labeled as 'U'. You can adjust the position and size of the lightning bolt by modifying these parameters.
\end{tikzpicture}
\end{figure}
%\end{document}
}

Motivated by the above, in this paper
we propose a test-time approach for optimally adapting to distribution shifts which arise due to changes in the underlying joint prior between the class labels $y$ and the nuisance labels $z$. 
We can view these changes as due to a hidden common cause $u$, such as the location of a specific hospital.
Thus we assume $\psource(u) \neq \ptarget(u)$,
where $\psource$ is the source distribution,
and $\ptarget$ is the target distribution.
Consequently,
$p_i(y,z) =\sum_u p(y,z|u) p_i(u)$ will change
across domains $i$.
However, 
we assume that the generative model of the features
is invariant  across domains,
so $p_i(\vx \mid y, z) =p(\vx \mid y, z)$.
%Although this is a strong assumption, 
%it becomes reasonable if $z$
%contains all the relevant ``factors of variation''.
See Figure~\ref{fig:model} for an illustration of our modeling assumptions.

The key observation behind our method is that 
our assumptions are equivalent to the standard
 ``label shift assumption'',
 except it is defined
 with respect to an expanded label $m=(y,z)$, which we call the meta-label. 
 We call this the ``expanded label shift assumption''.
 This lets use existing label shift techniques,
such as 
\citet{Alexandari2020,Lipton2018,Garg2020shift},
to adapt our model using 
a small sample of \textit{unlabeled} data $\{ \vx_n \sim p_t(\vx) \}$
from the target domain
to adjust for the shift in the prior over meta-labels,
as we discuss in \cref{sec:EM}.
Importantly, although our approach relies on the assumption that  $p(\vx \mid y, z)$ is preserved across distribution shifts, it is based on learning a {\em discriminative} base model $p_s(y, z, \mid \vx)$,
which we adjust
to the target distribution $p_t(y \mid \vx)$,
as we explain 
in \cref{sec:likelihood}.
Thus we do not need to fit a generative model to the data.
We do need access to labeled examples
of the confounding factor $z$ at training time, but such data is often  collected anyway
(albeit in limited quantities) especially for protected attributes.
Additionally, because it operates at test-time, our method does not require retraining to adapt the base model to multiple target domains.
We therefore 
call our approach Test-Time Label Shift Adaptation (TTLSA).

We evaluate TTLSA on  various standard image and text classification benchmarks,
as well as the \chexpert chest X-ray dataset.
%two experimental settings that have been used before to demonstrate the effect of spurious correlations in image data: \cmnist~\citep{arjovsky2019invariant,gulrajani2021in} and \chexpert~\citep{chexpert}.
We show that in shifted target domains TTLSA outperforms ERM (empirical risk minimization, which often uses spurious features inappropriately in target domains),
as well as methods that  train invariant classifiers (which ignore spurious features).
Note, however,  that it has been shown that no single adaptation method can work under all forms of distribution shift \citep{Veitch2021,Kaur2022}.
Our assumptions capture certain kinds of shift, but certainly not all.
In particular, our method is unlikely to help with the kinds of problems studied in the domain adaptation literature,
where there is covariate shift (i.e., a change from
$\psource(\vx)$ to $\ptarget(\vx)$) which is not captured
by a change in the distribution over the
causal factors $(y,z)$.

\section{Related work}
\label{sec:related}

% Jian2022
\paragraph{Spurious correlations, invariant learning, and worst-group performance}
Spurious correlations have mostly been studied in the
domain generalization literature
(see e.g., \citep{wilds,gulrajani2021in}),
in which a  model is expected to generalize (i.e., achieve acceptable performance) in a target domain without access to any data from that target domain.
In this problem setup,  building predictors based on  the principle of invariance or minimax optimality with respect to spurious correlation shifts is a natural approach.

Many of these methods make the same
modeling assumptions as we do 
(shown in Figure~\ref{fig:model});
this  is often referred to as an anti-causal prediction setting,
since the   labels cause (generate) the features rather than vice versa
\citep{Schoelkopf2012,Veitch2021, Makar2022, Puli2022, Zheng2022shortcut}.
These methods use the fact that an invariant predictor will satisfy certain conditional independencies, and employ regularizers that encourage the desired conditional independence.
A related line of work tries to minimize the worst-case loss across groups (values of $m=(y,z)$) using distributionally robust optimization \citep[see, e.g.,][]{Sagawa2020,JTT,Nam2022ssa,Lokhande2022}).
Recently, \citet{idrissi2022simple} showed empirically that
the ``SUBG'' method, which uses ERM on a group-balanced version of the data achieved by subsampling, can be an effective approach to learning worst-group-robust predictors in this setting.

\eat{
The finding is corroborated by theory in \citet{Makar2022}, under a special case of Figure~\ref{fig:model} that \citet{Veitch2021} call the ``purely spurious'' case.
Specifically, when the portions of $X$ can be fully partitioned into pieces that are exclusively causally influenced by either $Y$ and $Z$, the optimal invariant predictor corresponds to the Bayes optimal predictor in an ideal, balanced distribution.
\citet{Makar2022fairness} argue that this assumption is often reasonable in the context of medical imaging data, consistent with our experimental results.
}

\paragraph{Unsupervised domain adaptation and label shift}
In the Unsupervised Domain Adaptation (UDA)  literature,
we assume that unlabeled data from the target domain is available, either at training or test time.
To make optimality guarantees, UDA methods require assumptions about the structure of the distribution shift.
Methods are often categorized into whether they require a covariate shift assumption
(i.e., that the discriminative distribution $p(y \mid \vx)$ is preserved \citep[see, e.g.,][]{shimodaira2000improving}),
or a label shift assumption
(i.e., that generative distribution $p(\vx \mid y)$ is preserved \citep{Lipton2018,Garg2020shift}).

In the setting of Figure~\ref{fig:model}, neither the standard covariate shift nor label shift assumptions hold; however, the label shift assumption does hold with respect to the meta-label $m=(y, z)$.
Our primary contribution is to show that label shift adaptation techniques, 
such as \citet{Alexandari2020},
when 
applied with respect to the meta-label, can also be used to adapt to changes due to spurious correlations.

Interestingly, in the ``purely spurious'' setting described in \citet{Makar2022}, which is a special case of Figure~\ref{fig:model}, we can obtain a risk-invariant model as a special case, by adapting to a target distribution for which $\ptarget(y, z) = \psource(y)\psource(z)$ \citep{Veitch2021,Makar2022fairness}.
\footnote{\label{foot:purely spurious}The ``purely spurious'' term was coined in \citet{Veitch2021}, but for brevity we describe using formalism from \citet{Makar2022}.
Within the model of Figure~\ref{fig:model}, an association is ``purely spurious'' if there is a sufficient statistic of $X$, $e(X)$, such that $\indep{Y}{X} \mid e(X), Z$ and $\indep{e(X)}{Z} \mid Y$.
This occurs when the influence of $Y$ on $X$ can be localized in some features $e(X)$ that are not affected by $Z$, or intuitively, when the influences of $Y$ and $Z$ on $X$ are disentangleable.
The results of \citet{idrissi2022simple}, where spurious correlations are neutralized by data balancing, suggests that pure spuriousness is a common case, at least among worst-group benchmarks.
We discuss pure spuriousness, worst-group performance, and invariance in more detail in the supplement.}
Based on this observation, we show that logit adjustment \citep{Menon2021}, a set of test- and training-time methods developed for long-tail learning, can be repurposed to do invariant learning when applied to the meta-label $m=(y,z)$.

%UDA methods attempt to reduce the negative effect of distribution shifts by training a joint model on the labeled source and the unlabeled target data.
%Some prior works do so using feature alignment techniques~\citep{peng2019moment,dan,ddc} or reconstructing source and target data using autoencoders~\citep{drcn} or cycle-consistency GANs~\citep{cycada}. Adversarial learning based UDA methods also use adversarial two-player games to disentangle domain invariant and domain specific features~\citep{dann,cdan,dirtt}. Self-training based methods use the source domain classifier to create pseudo labels for adaptation~\citep{mei2020instance,zhang2021prototypical,daformer}. 
%Note that all UDA approaches need to access both labeled source
%data and unlabeled target data during training, which is a special case for the more challenging setting of test-time adaptation discussed below.

\paragraph{Methods using the expanded label shift assumption}
The ``generative multi-task learning'' or GMTL method of
\citet{Makino2022}
makes the same expanded label shift assumption that we do.
However, instead of estimating $\ptarget(y,z)$ from unsupervised target data, they instead assume that there
is some value $\alpha$ 
such that $\ptarget(y,z) = \psource(y,z)^{1-\alpha}$.
They state that choosing $\alpha$ is an open problem,
and therefore they report results for a range of possible $\alpha$'s.
%(Using $\alpha=0$ recovers a standard unadapted ERM model.)
By contrast, we provide a way to estimate $\ptarget(y,z)$,
and we do not restrict it to have the above functional form.
%Furthermore, we show the importance of having a calibrated base classifier when using these kind of reweighting techniques.
%We show that this gives inferior results compared to our approach, even with the optimal choice of $\alpha$.
\eat{
They also assume the conditionally factorized distribution
$p(y,z|\vx) = p(y|\vx) p(z|\vx)$,
so that they can use existing multi-task learning methods;
however, this is not strictly necessary,
since we can always combine $(z,y)=m$ into a single ``mega label'',
albeit at the cost of increasing the number of model parameters.
Finally, instead of computing the marginal
distribution $\ptarget(y|\vx)  = \sum_z \ptarget(y,z|\vx)$,
they pick the maximum of the joint distribution,
$\hat{y},\hat{z} = \arg \max \ptarget(y,z|\vx)$,
although this is another unnecessary simplification.
}

%We discuss GMTL in greater detail in Section~\ref{sec:GMTL}.
In \citet{Jiang2022},
they propose a method called Anti-Causal Transportable and Invariant Representation or ``ACTIR''
which also makes the same expanded label shift assumption that we do,
and further relaxes the assumption that $Z$
 is observed during source training. 
 However, they require access to examples from multiple source distributions, 
which they use to  learn a domain invariant classifier.
Furthermore, then require labeled $(x,y)$ examples
to adapt their classifier at test time.

\paragraph{Test time adaptation}
There is a growing Test Time Adaptation (TTA) literature that explores strategies for adapting a trained model at test time using unlabeled data from the target domain.
These methods are based on various heuristics to adapt discriminative models without specifying strong assumptions on the distribution shift structure.
For example,
TENT~\citep{tent}  uses entropy minimization to update the batch normalization layers of a CNN,
and MEMO~\citep{memo} uses ensembles of predictions for different augmentations of a test sample.
\eat{
However, this is prone to error accumulation when predictions are miscalibrated,
and is restricted to particular DNN architectures.
SHOT~\citep{shot} uses clustering-based pseudo labeling along with information maximization. MEMO~\citep{memo} uses ensembles of predictions for different augmentations of a test sample. AdaContrast~\citep{adacontrast} utilizes contrastive self-supervised learning jointly with online pseudo labeling.  
%Our method is an example of 
%``just-in-time unsupervised domain adaptation''
% \citep{Rosenfeld2022}.
}
Most of the TTA literature has focused on models that work on images (e.g., by leveraging data augmentation), 
so these techniques cannot be applied to embeddings or other forms of input. By contrast,  our method can be applied to any classifier,
 even  non-neural ones, such as random forests.
For a more  comprehensive review of TTA methods,
see \citep{liang2023ttasurvey}.

\section{Test-Time Label Shift Adaptation}

Our aim is to 
construct a Bayes optimal predictor for the target distribution, $f_t(\vx)$, using
a large labeled dataset from the source distribution,
 $\data_s^{xyz} = \{(\vx_n, y_n, z_n) \sim \psource(\vx, y, z)\}$,
 and a small unlabeled dataset from the target distribution,
 $\data_t^{\vx} = \{\vx_n \sim \ptarget(\vx)\}$.
The optimal prediction for the class label in the target domain is given by 
\begin{equation}
\begin{aligned}
f_t(\vx) =& \arg\max_y \ptarget(y|\vx) 
= \arg\max_y \sum_z \ptarget(y, z|\vx).
\end{aligned}
\label{eqn:MAP}
\end{equation}
where the joint posterior over the class label $y$
and the nuisance factor $z$ is given by
\begin{equation}
\label{eq:posterior}
\begin{aligned}
\ptarget(y,z|\vx) \propto
\ptarget(\vx|y,z) \ptarget(y,z) 
= \psource(\vx|y,z) \ptarget(y,z)  
%\propto \frac{\psource(y,z|\vx)}{\psource(y,z)} \ptarget(y,z).
\end{aligned}
\end{equation}
where the first step follows from Bayes' rule,
and the second step follows from our causal stability
assumption that $\ptarget(\vx|y,z) = \psource(\vx|y,z) = p(\vx|y,z)$.

The first key insight of our method is that we can use the EM algorithm
(or other optimization methods) to estimate $\ptarget(y,z)$
by maximizing the likelihood of 
the unlabeled target dataset, as explained in \cref{sec:EM}.
The second key insight is that we can estimate the likelihood 
$p(\vx|y,z)$ up to a constant of proportionality by 
fitting a discriminative model on the source dataset,
and then dividing out by the source prior:
\begin{align}
  \psource(\vx|y,z) \propto
  \frac{\psource(y,z|\vx)}{\psource(y,z)}
\label{eqn:scaledLik}
\end{align}
This is known as the ``scaled likelihood trick'' \citep{Renals1994},
and avoids us having to fit a generative model.
We can estimate $\psource(y,z|\vx)$ using any supervised learning method.
However, to get good performance in practice,
we need to ensure the probabilities are calibrated, as we explain in
\cref{sec:likelihood}.
Finally, we can estimate the source prior $\psource(y,z)$ just by counting how often each $(y,z)$ combination occurs 
in  $\data_s^{xyz}$
and then normalizing.
Combining \cref{eq:posterior} with \cref{eqn:scaledLik}
we can compute the posterior distribution over augmented labels
from the unlabeled target data as follows:
\begin{equation}
\label{eq:posteriorE2E}
\begin{aligned}
\ptarget(y,z|\vx) \propto
\frac{\psource(y,z|\vx)}{\psource(y,z)} \ptarget(y,z).
\end{aligned}
\end{equation}
We can then compute $f_t(\vx)$ using \eqref{eqn:MAP}.
In summary,  our TTLSA method consists of the following two steps:
\begin{enumerate}
\item \textbf{Train on source.} Train a model $\psource(y,z|\vx)$ using supervised learning
(and calibration)
applied to $\data_s^{xyz}$,
as explained in \cref{sec:likelihood}.
Also estimate $\psource(y,z)$
from $\data_s^{xyz}$
by counting.

\item \textbf{Adapt to target.} Estimate $\ptarget(y,z)$ using EM applied to $\data_t^{x}$,
as explained in \cref{sec:EM}.
Then compute $\ptarget(y,z|\vx)$ using \eqref{eq:posterior}
and  $f_t(\vx)$ using \eqref{eqn:MAP}.
\end{enumerate}

We discuss these steps in more detail below.

%We can estimate $\psource(y,z)$ empirically,
%since we assume we observe the meta-labels $m=(y,z)$ at training time. Similarly, we can fit $\psource(m|\vx)$ using any kind of discriminative classifier, such as a neural network or random forest. The only remaining challenge is estimate $\ptarget(y,z)$ from unlabeled samples from $\ptarget(\vx)$.
%We discuss how to do this below.

\subsection{Fit model to source distribution}
\label{sec:likelihood}

In the first step, we fit a discriminative classifier 
 $\psource(y,z|\vx)$ on the source dataset.
This is just like a standard classification problem,
except we use an augmented  output space,
consisting of the class label $y$ and group label $z$;
we denote this joint label by $m:=(y,z)$.

\paragraph{Calibration}
\label{sec:calibration}

Our adaptation procedure crucially relies on the fact that the output of the classifier, $\psource(m|\vx)$, can be interpreted as calibrated probabilities.
Since modern neural networks are often uncalibrated
\citep[see, e.g.,][]{Guo2017-wm},
we have found it important to perform an explicit calibration step.
(See the supplementary for an ablation study where we omit the calibration step.)
Specifically, we follow 
 \citet{Alexandari2020} and use their  ``bias corrected temperature scaling'' (BCTS) method, which is a generalization of Platt scaling to the multi-class case.
In particular, let $l(\vx)$ be the vector of $M$ logits.
We then modify $\psource(m|\vx)$ as follows:
\begin{equation}
\psource(m |\vx) \propto \exp\left(\frac{l(\vx)_m}{T} + b_m\right)
\end{equation}
where $T \geq 0$ is a learned temperature parameter, and $b_m$ is a learned bias.
This calibration is done after the source classifier is trained
using a labeled validation set from the source distribution.

 \paragraph{Logit adjustment}
\label{sec:logit}

The calibration method  is a post-hoc method.
However, if the source distribution has a 
``shortcut'', based on spurious correlations between $z$ and $y$, the discriminative model may exploit this by overfitting to it.
The resulting model will not learn robust features,
and will therefore be hard to adapt,
even if we use calibration.
To tackle this,
we take inspiration from Proposition 1 of
\citet{Makar2022},
which showed that a classifier that is trained on the unconfounded or balanced distribution $\pbalanced$, such that $\pbalanced(y,z) = p_s(y) p_s(z)$, will not learn any ``shortcut''
between the confounding factor $z$ and
the target label $y$.
Such a model will therefore have a risk which is invariant
to changes in the $p(z|y)$ distribution, and, in ``purely spurious'' anti-causal settings (a special case of Figure~\ref{fig:model}, see footnote~\ref{foot:purely spurious}), they showed that this is indeed the optimal one among all risk-invariant predictors (see discussion in \cref{sec:related}).
 If both marginals are uniform, then
 $\pbalanced(y,z) \propto 1$,
 so the corresponding balanced version of the source distribution becomes
$
\pbalanced(m|\vx) \propto \frac{\psource(m|\vx)}{\psource(m)}
$,
which in log space becomes
$\log(\pbalanced(m|\vx)) \equiv
\log(\psource(m|\vx)) - \log(\psource(m))$.
We train our classifier to maximize this
balanced objective;
this is known as ``logit adjustment''
\citep{Menon2021}.
After fitting, we can recover the  classifier 
for the original source distribution
using
$\psource(y,z|\vx) \propto \pbalanced(y,z|\vx) \psource(y,z)$.
%assuming $\pbalanced(y,z)=\psource(y) \psource(z)$.
Finally we apply calibration to $\psource(y,z|\vx)$,
 as explained above.
(Note that when the marginals are not uniform, one can apply a similar adjustment, subtracting off $\log(\psource(y,z) / (\psource(y)\psource(z))$.)

%\todoqingyao{We should mention TTSLA on Logit Adjustment works better than TTSLA on Unadapted.}

\eat{
\paragraph{Learning with fewer group labels}
If only a small set of group labels are only  available
(e.g., as part of a validation set),
then we can use semi-supervised learning methods
to fit  $\psource(y,z|\vx)$.
In this paper, we focus on the fully supervised setting,
although we have conducted preliminary experiments in the semi-supervised setting as well.\footnote{
In particular, we fit $\psource(y,z|\vx)$ on a small fully
labeled subset, $\data_s^{xyz}$,
and then imputed the missing $z$ values
on the larger $\data_s^{xy}$
by applying this model and only keeping the most confident predictions.
We then refit the classifier on this larger imputed dataset
and used it for TTLSA.
We got good results even when 90\% of the $z$ labels were missing.
However, we leave more detailed evaluation of this semi-supervised extension to future work.
 }  %
 }

\subsection{Adapt model to target distribution}
\label{sec:targetPrior}
\label{sec:EM}

Given some unlabeled samples from the target distribution,
$\data_t^x = \{ \vx_n \sim \ptarget(\vx): n=1:N \}$,
our goal is to estimate the new prior, 
$\ptarget(y,z)=\ptarget(m) = \pi_m$.
This is a standard subroutine in label shift adaptation methods \citep{Lipton2018,Garg2020shift}; our innovation is to do this with respect to the meta-label $m$.
Here, we use an EM approach \citep{Alexandari2020}, and maximize the log likelihood
of the unlabeled data,
$\mathcal{L}(\vx; \vpi, \vtheta) = 
\sum_{\vx_n} \log \ptarget(\vx_n; \vpi, \vtheta)$,
wrt $\vpi$,
where 
\begin{equation}
\begin{aligned}
\ptarget(\vx; \vpi, \vtheta)
= \log\left(\sum_m 
\pi_m \ptarget(\vx|m; \vtheta) \right) 
=  \log\left(\sum_m \pi_m \frac{\psource(m|\vx; \vtheta)}{\psource(m;\vtheta)} \right)
+ \text{const}
\end{aligned}
\end{equation}
where $\vtheta$ are the parameters estimated from the source
distribution.
This objective is a sum of logs of a linear function of $\vpi$,
and is maximized subject to the affine
constraints $\pi_m \geq 0$ and $\sum_{m=1}^M \pi_m=1$.
Thus, under weak conditions on the ratios ${\psource(m|\vx; \vtheta)}/{\psource(m;\vtheta)}$ (these implied when $\vx$ contains some information that can discriminate between levels of $m$) the problem is concave,
with a unique global optimum,
implying the parameters are identifiable \citep{Alexandari2020}.

A simple  way to compute this optimum is to use EM,
which automatically satisfies the constraints.
Let $\pi_m^j$ be the estimate of $\pi_m$ at iteration $j$.
We initialize with $\pi_m^0 = \psource(m)$ and run the following iterative procedure below:
\begin{align}
\ptarget(m|\vx_n; \vpi^j, \vtheta) &\propto 
\ptarget(\vx_n|m; \vtheta) \pi_m^j
\propto \frac{p_s(m|\vx; \vtheta)}{p_s(m; \vtheta)} \pi_m^j
 \text{ // E step } \\
\pi_m^{j+1} &=
\frac{1}{N} \sum_{n=1}^N
\ptarget(m|\vx_n; \vpi^j, \vtheta)
\text{ // M step }
\end{align}
%This converges within $J \sim 10$ terations to the global maximum.
We then set $\ptarget(m) = \vpi_m^J$.
We can also modify this to compute a MAP estimate, instead of the MLE, by using a Dirichlet prior.
See 
%Section~\ref{sec:em-deriv} 
the supplementary 
for a detailed derivation.

After estimating $\ptarget(y,z)$ on $\data_t^x$, we can compute
$\ptarget(y,z|\vx)$ for the examples in $\data_t^x$
using \cref{eq:posteriorE2E}.

\section{Experiments}

In this section, we provide an experimental comparison of our method with various baseline methods on a variety of datasets.
In particular, we compare the following methods:
\begin{description}

\item[ERM] This corresponds to training a model on the source distribution, and then applying the same model to the target distribution without any adaptation, i.e., we assume $\ptarget(y|\vx) = \psource(y|\vx)$.

\item[gDRO] The group DRO method
of \citep{Sagawa2020} is designed to optimize
the performance of the worst performing group.
(We only use this method for the worst-group benchmarks in \cref{sec:facebook}.)

\item[SUBG]  The  ``SUBG'' method of 
 \citep{idrissi2022simple}  subsamples the data so there is an equal number of examples in each group $m=(y,z)$, then trains a classifier by standard ERM. 
 This is a simpler alternative to gDRO yet often achieves comparable performance.
 %Despite its simplicity, this method yields SOTA results on several worst-group benchmarks.

\item[LA] This is our logit adjustment method of \cref{sec:logit}, which  approximates a domain invariant classifier by targeting a uniform prior on the source domain, thus avoid overfitting to spurious correlations.

\item [TTLSA] This is our  EM method of \cref{sec:EM}
that adapts the LA-based classifier using unlabeled data.
%We consider $N \in \{64, 512\}$.

\item[Oracle] This is similar to TTLSA, but we replace the EM procedure with the ground truth  target meta-label prior $\ptarget(y,z)$. 
This gives an upper bound on performance.
(We only use this for the CMNIST experiments
in \cref{sec:cmnist}
and the \chexpert experiments in \cref{sec:chexpert}, 
where we artificially control the degree of distribution shift.)

\eat{
\item[GMTL]  This is the GMTL baseline we discussed in \cref{sec:GMTL}.
We  use $\alpha=1$,
which targets an invariant classifier.
%For values of $\alpha$ in between 0 and 1,
%the results lie between the unadapted model and the invariant model.
%, as we show in the appendix.
}

\end{description}

%We show that our method  comes close to the oracle performance, and outperforms Unadapted, GMTL, and Invariant.

\begin{figure*}
\centering
\begin{subfigure}[b]{0.45\textwidth}
\centering
 \includegraphics[width=0.95\textwidth]{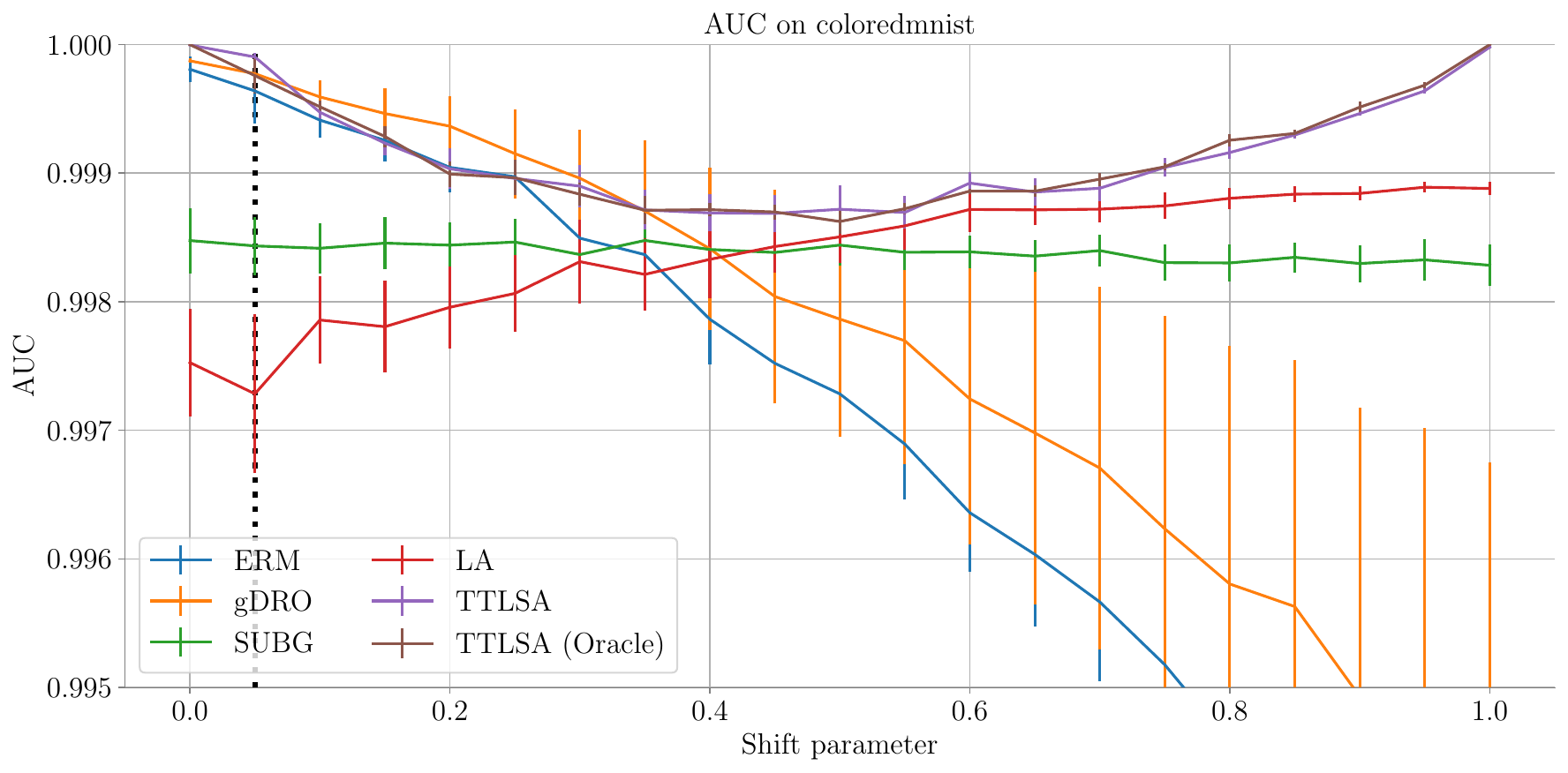}
\caption{ }
        \label{fig:mnist-noise0-domain1-cali}
\end{subfigure}
\begin{subfigure}[b]{0.45\textwidth}
\centering
\includegraphics[width=0.95\textwidth]{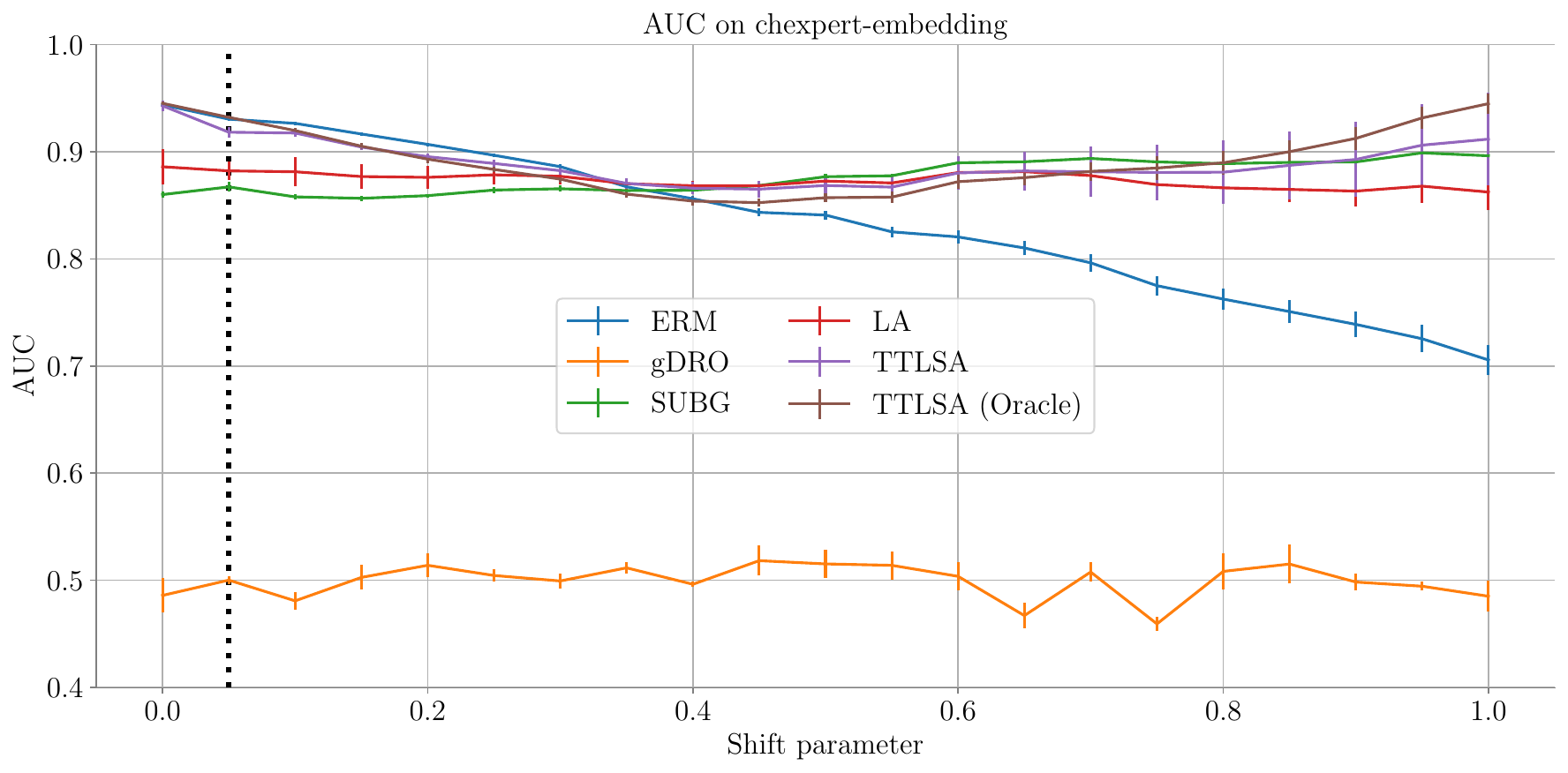}
\caption{ }
\label{fig:chexpert-embedding-domain1-cali}
\end{subfigure}
 \caption{Test set AUC performance on  (a) \cmnist and (b) \chexpert  as we vary the  correlation between $y$ and $z$ in the target distribution. The vertical dotted line marks the source distribution. There are three performance regimes: (1) the unadapted ERM model (blue line) degrades smoothly under shift; (2) the invariant models (yellow, green, and red lines) have flat performance; and (3) the adapted models (purple and brown lines) out-perform invariant models, yielding a U-shape.
 }
\end{figure*}

\subsection{\cmnist}
\label{sec:cmnist}

In this section,  we apply our method
to the \cmnist dataset \citep{arjovsky2019invariant,gulrajani2021in}.

\paragraph{Dataset}

\begin{table}
\centering
\begin{adjustbox}{width=0.5\textwidth}
% \centerline{
\begin{tabular}{cc|cc}
& & \multicolumn{2}{c}{z}  \\
& & 0 & 1 \\ \hline
\multirow{2}{*}{y} & 0 & 0.5 & 0  \\
& 1 & 0 & 0.5
\end{tabular}
\hspace{1cm}
\begin{tabular}{cc|cc}
& & \multicolumn{2}{c}{z}  \\
& & 0 & 1 \\ \hline
\multirow{2}{*}{y} & 0 & 0 & 0.5  \\
& 1 & 0.5 & 0
\end{tabular}
% }
\end{adjustbox}
\caption{
The two ``anchor'' distributions, reflecting total positive and negative correlation between the class label $y$ and the confounding factor $z$.
Left: $p_0$.
Right: $p_1$.
From these distributions, we can create a family of target distributions $p_{\lambda} = \lambda p_0 + (1-\lambda) p_1$.
}
\label{tab:anchors}
\end{table}

We construct the dataset in a manner similar to \citep{arjovsky2019invariant,gulrajani2021in}.
Specifically, we create a binary classification problem, where label $y=0$ corresponds to digits 0-4, and $y=1$ corresponds to 5-9.
We then sample a random color $z \in \{0,1\}$,
corresponding to red or green,
for each image,
with a probability that depends on the target distribution.
See 
the supplementary material 
%Figure~\ref{fig:mnist-sample}
for a visualizaiton of this dataset.
Since the color is easier to recognize than the shape, color can act as a ``shortcut'' to predicting the class label, even though this is not a robust (domain invariant) feature.

\eat{
We create 21 distinct target distributions,
each with a correlation between $y$ and $z$
given by $\rho_i = \frac{1}{20} i$.
So the joint prior distribution over the meta-labels is given by
\begin{equation}
p_{i}(y, z) = \begin{array}{c|cc}
_{\large y}\backslash^{\large z} & \text{0 (red)} & \text{1 (green)} \\
\hline
0 & \frac{1}{2}(1 - \rho_i) & \frac{1}{2}\rho_i \\
1 & \frac{1}{2}\rho_i & \frac{1}{2}(1 - \rho_i) \\
\end{array} 
\end{equation}
Once we sample $m=(z,y)$, we then pick a random image from the training set with label $y$,
and then apply the color specified by $z$.
}

We create a set of 21 target distributions
$p_{\lambda} = (1-\lambda) p_0 + \lambda p_1$,
where $\lambda \in \{0, 0.05, \ldots, 0.95, 1.0 \}$,
and $p_0(y,z)$ and $p_1(y,z)$ are two anchor distributions
in which $y$ and $z$ are correlated and anti-correlated,
respectively
(see \cref{tab:anchors}).
By changing $\lambda$, we can control the dependence between $y$ and $z$.
We train the classifier on a source domain exhibiting a strong spurious correlation
(we choose $\lambda=0.05$),
and then apply the model
to each target domain,
$p_{\lambda}$,
for $\lambda \in \{0, 0.05, \ldots, 0.95, 1.0 \}$.
We measure performance using Area Under the ROC curve (AUC).
\eat{
this is a better metric than classification accuracy,
since many of the distributions 
have  a very skewed label prior,
%(see e.g. Table~\ref{tab:chexpert-baseline}),
which means the base rate accuracy can already be very high.
}

\paragraph{Training procedure}

During training, we fit a LeNet CNN on the training set by using $m=(y,z)$ as the label.
We use AdamW with a batch size of 64 and a learning rate of $10^{-3}$ for a maximum of 5000 epochs, and run calibration for a maximum of 1000 epochs with the same learning rate and batch size on a 10\% holdout set.
However, training typically terminates much earlier since we also calculate the exponential moving average of the validation loss with a decay rate of 0.1, and stop early when the smoothed validation loss does not decrease for 5 consecutive epochs.
%We also fit a gradient boosted classifier to the pixels.
%
During evaluation, we infer the target label prior for each target distribution using EM applied to an unlabeled test set
of size 64 or 512, and 
we then predict the class labels on this test set.
Finally we repeat this across all the unlabeled minibatches in the target distribution, and report the overall AUC.

\paragraph{Results}

\eat{
\begin{figure*}
    \centering
    \includegraphics[width=0.8\textwidth]{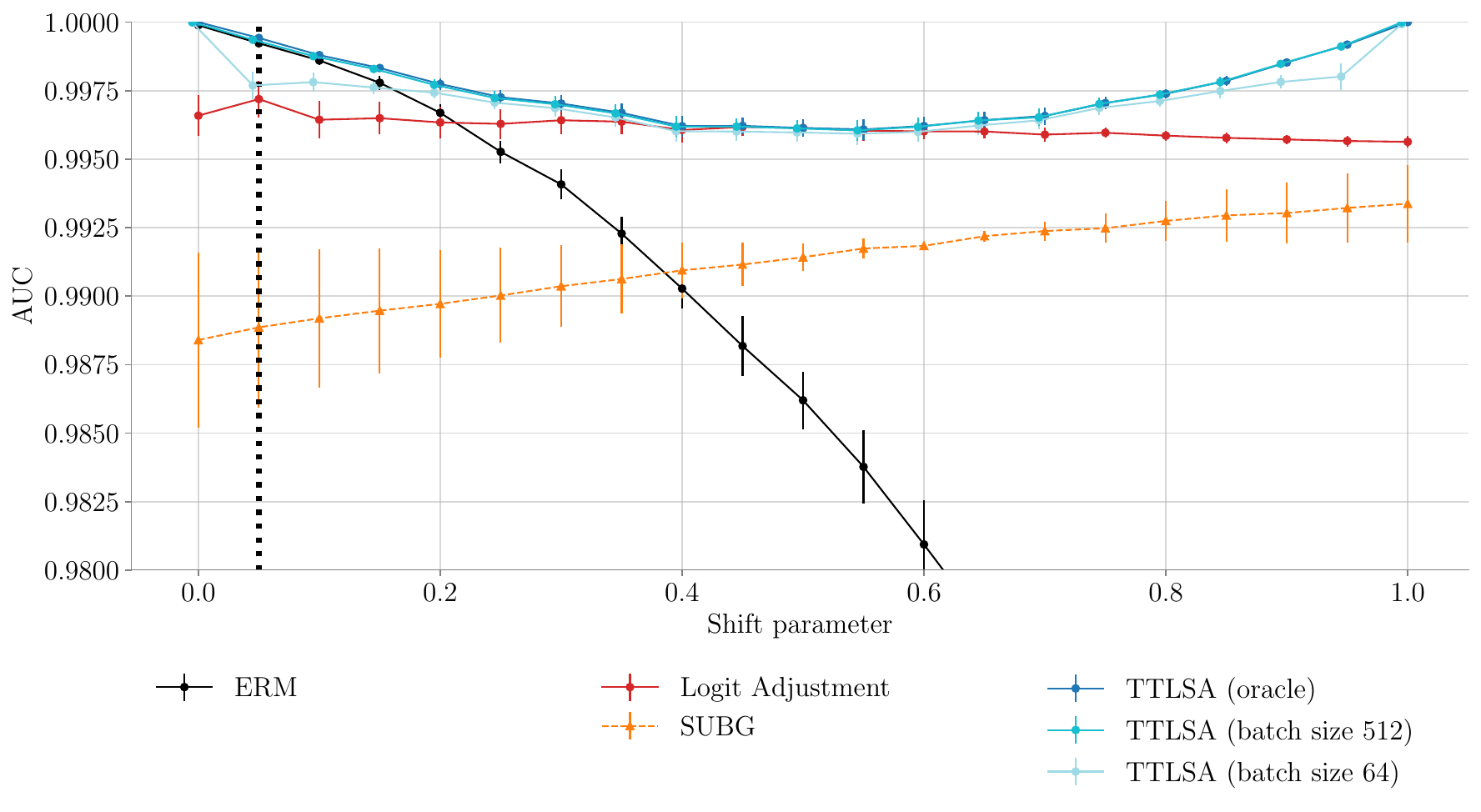}
    \caption{Performance across target distributions on \cmnist, where the shift parameter controls the marginal correlation between the digit label $y$ and the color $z$. The vertical dotted line marks the source distribution. There are three performance regimes: (1) the unadapted model degrades smoothly under shift; (2) the invariant models have flat performance; and (3) the adapted models out-perform invariant models in target distributions commensurate with the strength of prior information in the spurious correlation, yielding a U-shape.}
    \label{fig:mnist-noise0-domain1-cali}
\end{figure*}
}

\Cref{fig:mnist-noise0-domain1-cali}
shows our results,
with error bars showing
%standard deviation 
standard error of the mean across 4 trials. 
There are three main groups of curves.
(1) The ERM model (black line) shows performance that gets steadily worse as the target distribution shifts away from the source.
(2) The invariant models
(SUBG in dashed orange
and logit adjustment in solid red)
are approximately constant across domains,
as expected.
(In this experiment, we see that logit adjustment outperforms SUBG.)
3) Our adaptive TTLSA models (blue curves),
corresponding to TTLSA with
64 or 512 samples and the oracle method,
all show a U-shaped curve, which is an upper bound
on all the other curves.
In particular, we see 
that the U-shaped curve is tangent to the invariant line when the target distribution is unconfounded ($\lambda=0.5$); in that case, there is no prior information from $\ptarget(y,z)$ to exploit.
However, in all other target domains, the adapted prior information improves performance.
As we increase the size of the unlabeled target dataset,
we see that the performance of TTLSA approaches that of the oracle.
To illustrate the fact that our method can also be applied
to non-neural net classifiers
we also applied TTLSA  on top a  gradient boosted tree classifier.
The results  (shown in the supplementary)
are qualitatively similar
to those in \cref{fig:mnist-noise0-domain1-cali}.

\eat{
Here are the key takeaways:
\begin{itemize}
    \item Our EM algorithm closely matches the oracle method,
    with a larger adaptation set ($N=512$) giving better results.

    \item As in CheXpert,
    the GMTL  performance depends on $\alpha$,
    and is generally worse than our EM method.

    \item Surprisingly, the invariant curve (green line)
    in \cref{fig:mnist-result-confounded} is higher than
    the oracle (black line) when $\lambda$ is close to 0.5.
    Note, however, that these two classifiers are trained on different datasets (albeit of the same size, namely 28,000 images).
    In particular, the invariant model is trained on data samples from 
    $p_{0.5}(\vx,y,z)$,
    but all the other models (including the oracle)
    are trained on data sampled from $p_{0.1}(\vx,y,z)$.
    These differ not only in their label priors,
    but also in the images that are sampled.
    Since the invariant training set is more similar to the
    $\lambda=0.5$ test set than the oracle training set,
    it suggests that the model may simply be ``overfitting''.
\end{itemize}

}

\subsection{CheXpert}
\label{sec:chexpert}

Next,  we apply our method
to the problem of disease classification using chest X-rays
based on the CheXpert~\citep{chexpert} dataset.
Chest X-rays are a particularly relevant application area for our method because sensitive attributes, such as patient sex or self-reported race, can be readily predicted from chest X-rays \citep{gichoya2022ai}.
Recent work, e.g., \citet{Jabbour2020} and \citet{Makar2022}, has confirmed the potential for classifiers to exploit these features as spurious features.

\eat{
More recently, there has been some discussion about whether foundation model embeddings designed to support a diverse array of diagnostic tasks \citep[e.g.,][]{Sellergren2022} should encode these sensitive attributes.
In particular, \citet{Glocker2022} argues that these introduce a risk of bias.
Here, we present a set of results showing that, with appropriate adaptations, such information could be useful when models are applied to distinct target populations.
}

\paragraph{Dataset}

%https://stanfordmlgroup.github.io/competitions/chexpert/
%as used in other works on spurious
%correlations (e.g., \citep{Jabbour2020,Makar2022,Glocker2022}).
CheXpert has 224,316 chest radiographs of 65,240 patients.
See  the supplementary material 
for a visualizaiton of this dataset.
Each image is associated with
14 disease labels derived from radiology reports,
and 3 potentially confounding attributes (age, sex, and race),
as listed in the supplementary.
We binarized the attributes as in \citet{Jabbour2020},
taking age to be 0 if below the median and 1 if above,
and sex to be 0 if female and 1 if male.
As for the class labels,
we define class 0 as ``negative'' (corresponding to no evidence of a disease),
and class 1 as ``positive''  (representing the presence of a disease); images labeled ``uncertain'' for the attribute of interest are discarded.
Following \citep{Glocker2022}, we focus on
predicting the label $y$ = ``Pleural Effusion''
and use sex as the confounding variable $z$.
%which has an empirical prior of $p(y=0)=0.5753$ and $p(y=1)=0.4247$.
See Figure~\ref{fig:chexpert-sample} for some samples from the dataset.
For the input features $\vx$, we either work with the raw gray-scale images,
rescaled to size $224 \times 224$,
or we work with  1376-dim feature embeddings derived from the pretrained CXR model \citep{Sellergren2022}.
This embedding model was pre-trained on a large set of weakly labeled X-rays from the USA and India.
Note, however, that the pre-training dataset for CXR is distinct from the CheXpert dataset we use in our experiments.

% We use a version of the model provided by the authors that was trained using a CLIP loss, that improves upon the results the CXR Foundation authors got with supervised contrastive loss for their original paper.
% In particular, the embeddings are float vectors corresponding to the pooled last layer of EfficientNet-L2 for a model trained on CLIP loss.

% https://github.com/nalzok/label-shift/blob/9c7d0be361eefba3f13a4a940d33775f8b9affd4/tta/datasets/chexpert.py#L82

To compare performance under distribution shift, we 
created a set of 21 distributions,
$p_{\lambda} = (1-\lambda) p_0 + \lambda p_1$,
where $\lambda \in \{0, 0.05, \ldots, 0.95, 1.0 \}$
as before.
We train using $\lambda=0.05$, representing a strong spurious correlation at training time,
and test with all 21 values.
The test images are distinct from the training,
and each test distribution has 512 samples in total.
Each patient may have multiple images associated with them, but there is no patient overlap in the training and test distributions.

\paragraph{Training procedure}

When working with embeddings, we use a  linear logistic regression model,
following \citet{Sellergren2022}, due to its simplicity and its good performance. 
To train this, 
we use AdamW with a batch size of 64 and a learning rate of $10^{-3}$ for a maximum of 5000 epochs, and run calibration for a maximum of 1000 epochs with the same learning rate and batch size on a 10\% holdout set.
However, training typically terminates much earlier since we also calculate the exponential moving average of the validation loss with a decay rate of 0.1, and stop early when the smoothed validation loss does not decrease for 5 consecutive epochs.
When working with pixels, we use a CNN.
See  
%\cref{sec:chexpertPixels}
the supplementary for details.

\paragraph{Results}

\eat{
\begin{figure*}
\centering
\includegraphics[width=0.8\textwidth]{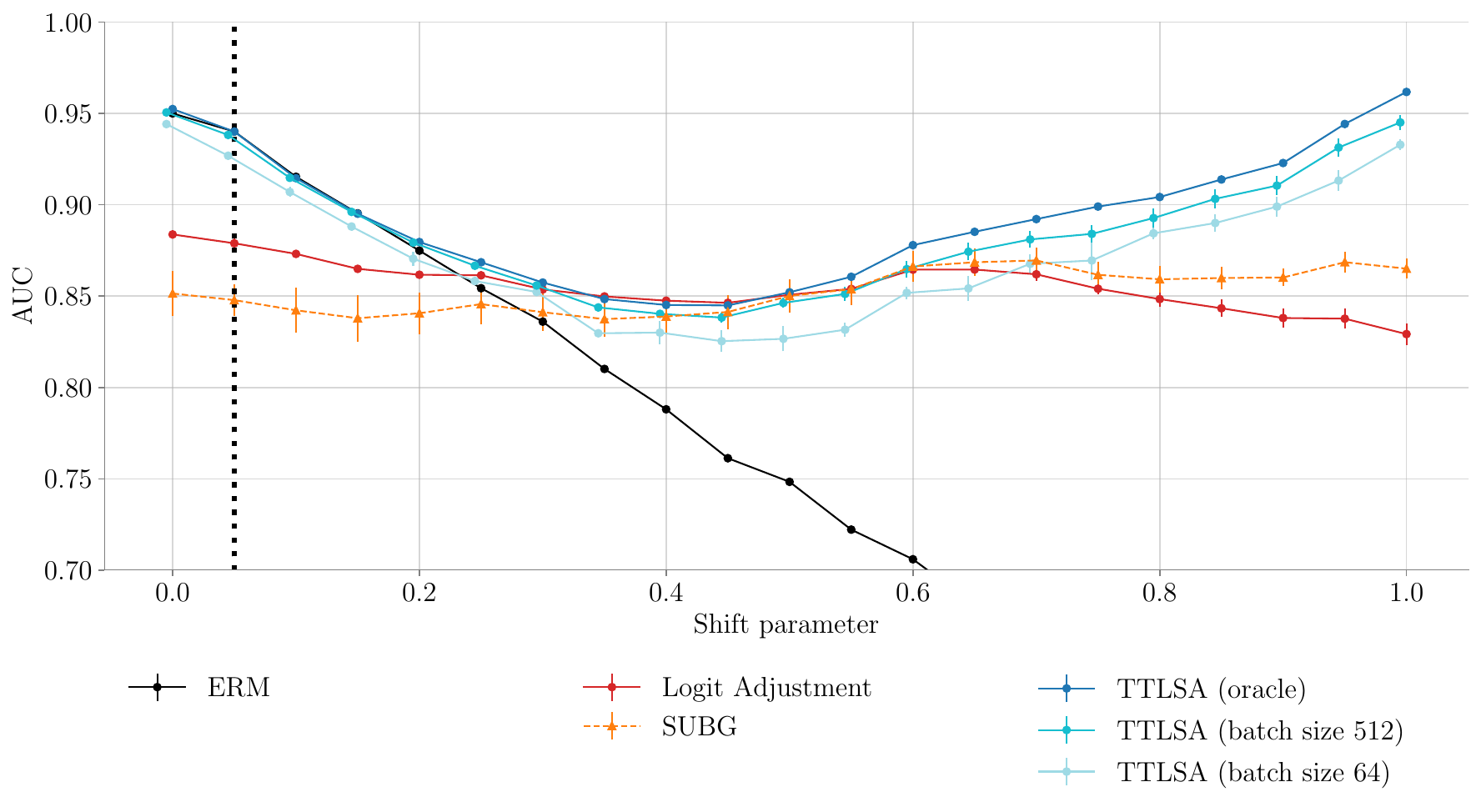}
\caption{
Performance across target domains on \chexpert embeddings, following the setup of Figure~\ref{fig:mnist-noise0-domain1-cali}.
}
\label{fig:chexpert-embedding-domain1-cali}
%\label{fig:chexpert-embedding-domain1}
\end{figure*}
}

\eat{
\begin{figure*}
\centering
\begin{subfigure}[b]{0.45\textwidth}
\centering
\includegraphics[width=\textwidth]{\figdir/chexpert-embedding-domain1-cali1000_auc_major.pdf}
\caption{ }
 \label{fig:chexpert-embedding-domain1-cali}
\end{subfigure}
\begin{subfigure}[b]{0.45\textwidth}
\centering
\includegraphics[width=\textwidth]{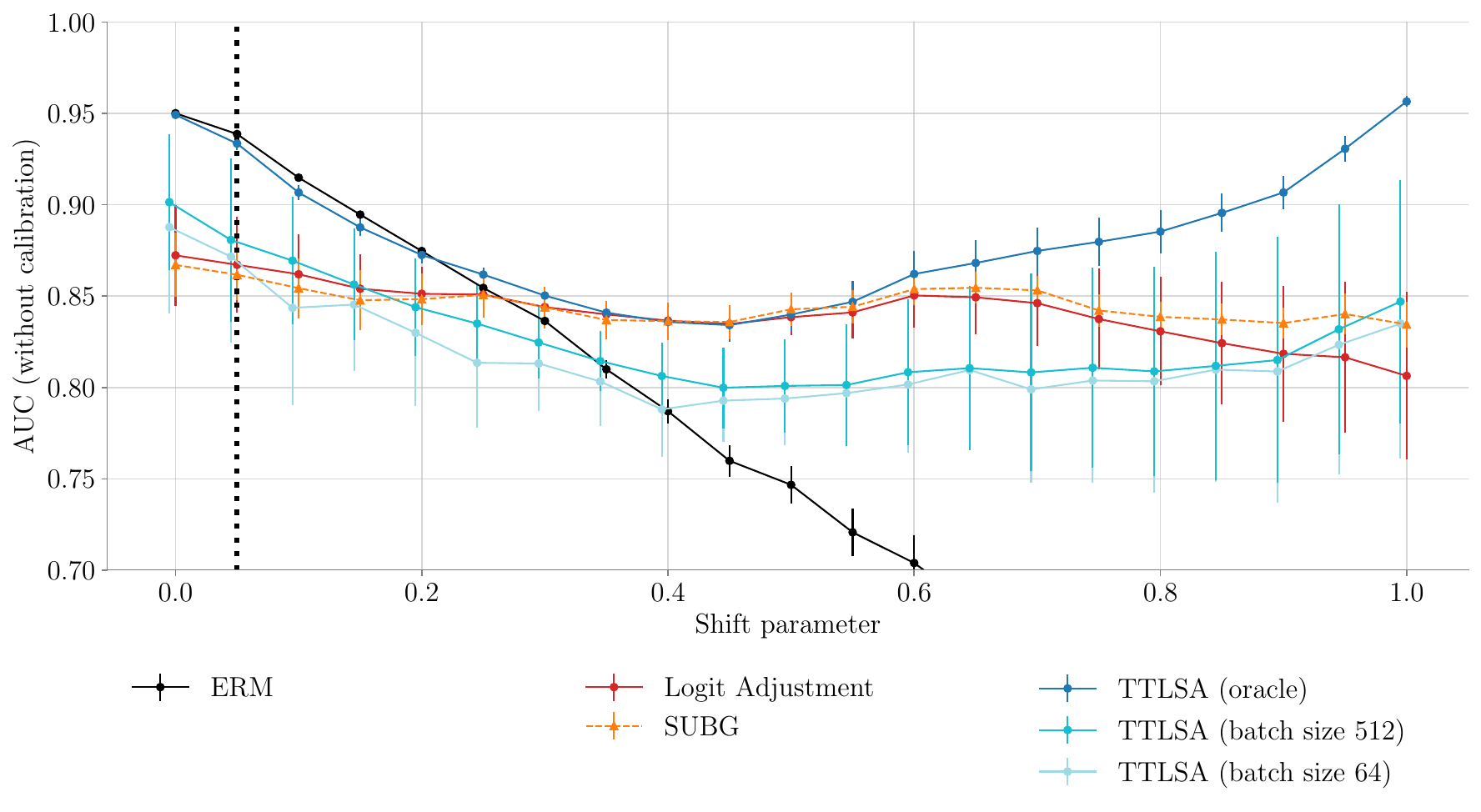}
\caption{ }
\label{fig:chexpert-embedding-domain1-nocali}
\end{subfigure}
\caption{
Performance across target domains on \chexpert embeddings, following the setup of Figure~\ref{fig:mnist-noise0-domain1-cali}.
(a) Results using calibration. Performance  mirrors those in Figure~\ref{fig:mnist-noise0-domain1-cali}.
(b) Results without calibration.
We see that calibration both improves performance and decreases variability between runs.
}
\label{fig:chexpert-embedding-domain1}
\end{figure*}
}

Our results for the embedding version of the data
are shown in \cref{fig:chexpert-embedding-domain1-cali}.
The trends are essentially the same as in the colored MNIST case.
In particular, we see that our TTLSA method outperforms the invariant baselines, and both adaptive and invariant methods
outperform ERM.
These results for the pixel version of the data are shown in 
\cref{sec:chexpertPixels}
%\cref{fig:chexpert-pixel-domain1}
in the supplementary.
We find that the relative performance of the methods is similar
to the embedding case,
but the absolute performance for all methods is 
 better when using embeddings, 
as was previously shown in  \citet{Sellergren2022}.

\paragraph{Discussion}

\citet{Glocker2022} point out that embeddings derived from X-ray classification models may contain information about sensitive attributes $z$, such as sex and age.
We confirmed this result, and were able to classify sex with an accuracy of 
over 95\% just using logistic regression on the CXR embeddings,
as shown in \cref{tab:chexpert-baseline}.
\citet{Glocker2022} argue that this may be harmful, since it can cause bias in the predictions of the primary label $y$ of interest (disease status).
As a counterpoint, our results also show that, if we can predict both $z$ and $y$ from the embeddings, this information can be used to make optimal adjustments for target populations featuring very different demographic makeups, in ways that can be beneficial for all groups.
However, our results also confirm such information does need to be handled with care.
%and then use $Z$ as an additional feature to predict $Y$,
%which can improve performance for all groups.

\subsection{Worst-group vision and text benchmark datasets}
\label{sec:facebook}

In this section, we apply our method to four benchmark datasets that
were first introduced in the group DRO paper
\citep{Sagawa2020},
and have since been widely used in the literature
on group robustness
\citep[see, e.g.][on data balancing]{idrissi2022simple}.
The four datasets are as follows.

\begin{description}
\item [CelebA] Introduced in \citep{celeba},
this is an image dataset of celebrity faces.
The class label $y$ is hair color (blond / not-blond)
and the group / attribute label $z$ is sex (male / female).
\eat{
The correlation
between $y$ and $z$
is mostly unchanged when moving from source
to target distributions.
}

\item [Waterbirds] Introduced in \citep{Wah2011,Sagawa2020},
this is an image dataset of birds
synthetically pasted onto two different kinds of backgrounds.
The class label $y$ is bird type (land bird or water bird),
and the group / attribute label $z$ is
background type
(land or water).
\eat{
The correlation
between $y$ and $z$
changes when moving from source
to target distributions in a way which makes relying
on this correlation harmful.
In particular, it changes from a strong correlation to a weak one, penalizing methods that take the shortcut of predicting $y$ based on features for $z$.
}

\item [MultNLI] Introduced in \citep{Williams2018},
this is a dataset of sentence pairs, $(s_1,s_2),$
where the goal is to predict if $s_1$ entails $s_2$.
The class label $y$ corresponds to entailment,
contradiction or neutral,
and the group / attribute label $z$ indicates
presence / absence of negation words.
\eat{
The correlation
between $y$ and $z$
changes when moving from source
to target distributions in a way which makes relying
on this correlation helpful.
In particular, it changes from a strong correlation to an even stronger one, rewarding methods that take the shortcut of predicting $y$ based on features for $z$).
}

\item [CivilComments (CC)] Introduced in \citep{Borkan2019},
This is a dataset of sentences from online forums.
The class label $y$ represents if the comment is toxic or not,
and the group/attribute label $z$ represents
whether the content is related to a minority group (such as LGBT) or not.
\eat{
The correlation
between $y$ and $z$
is mostly unchanged when moving from source
to target distributions.
}
\end{description}

\paragraph{Training procedure}
We use the code and hyper-parameters specified in 
\citet{idrissi2022simple} for the ERM, gDRO, and SUBG baselines.
We also use these same hyper-parameter values as ERM when fitting our
own model TTLSA, which is trained 
 to predict the augmented labels
$m=(y,z)$ using logit adjustment.
We run the experiments on sixteen Nvidia A100 40GB GPUs.
Depending on the dataset and method used, each experiment takes somewhere between 1 hour and 10 hours on an A100.
In total, the experiments took 415 GPU hours.

For each method, for worst-group accuracy evaluations, we tune the model using worst-group accuracy in a validation set, whereas for average accuracy evaluations, we tune the model using average validation set accuracy.
For worst-group evaluations, we use the LA baseline, without adjusting to the test set distribution.
This effectively targets a balanced group distribution, similar to SUBG.
For average target accuracy evaluations, we use the full TTLSA method with test-time EM adjustment.

\paragraph{Results}
We summarize our results in \cref{tab:facebook}.
We report the worst group and average group accuracy, averaged across 4 replication runs.
Our worst group numbers for the baseline methods
are within error bars for those
reported in  \citet{idrissi2022simple}.\footnote{
Our worst group accuracy results for CivilComments
are very different from those reported in \citet{idrissi2022simple}
despite using their code.
The reason is that they binarize the $z$ label during training, but use the fine-grained dataset with 9 possible $z$ values during evaluation.
%(see %\url{https://github.com/facebookresearch/BalancingGroups/issues/4}).
Our method requires that $z$ have the same set of possible values in train and test, so we use the coarse-grained dataset for both training and evaluation, and thus our performance metric is incomparable to theirs.
}
The \citet{idrissi2022simple} paper does not report average group performance, but we computed these results by modifying their code.

\begin{table}[h]
\centering
% \tiny
\begin{adjustbox}{width=\textwidth}
\begin{tabular}{l|c|c|c|c|c||c}
   Data & $m_s$ & $m_t$ & ERM  & gDRO & SUBG   & LA/TTLSA  \\
\hline
CelebA     &  0.44 & 0.49
& 80.83  (1.46) / {\bf 95.93} (0.03)
& {\bf 87.36} (0.47) / 94.68 (0.07)
& {\bf 87.10} (1.26) / 93.44 (0.19)
& 84.72 (0.58) / {\bf 95.55} (0.09)
\\
Waterbirds  & 0.73 & 0.39
& 85.78 (0.24) / 93.19 (0.16)
& 87.98 (0.86) / 93.06 (0.62)
& {\bf 88.87} (0.14) / 93.48 (0.11)
& {\bf 88.38} (0.36)  / {\bf 95.23} (0.34)
\\
MultiNLI  &  0.49 & 0.49
& 68.60 (0.40) / {\bf 82.70} (0.02)
& {\bf 76.79} (1.24) / 81.16 (0.07)
& 67.89 (0.91) / 72.15 (0.25)
& {\bf 76.33} (1.45) / {\bf 82.60} (0.04)
\\
CC  & 0.55 & 0.65
& 68.16 (1.03) / {\bf 88.00} (0.03)
& {\bf 79.66} (0.17) / 84.46 (0.43)
& 76.56 (0.25) / 79.56 (0.77)
& {\bf 79.27} (1.17) / 85.03 (0.71)
\\
\end{tabular}
\end{adjustbox}
\caption{
Accuracy of the worst / average $(y,z)$ group 
on the benchmark datasets.
%$\rho_s$ and $\rho_t$ are the empirical correlation coefficients between the label $y$ and the confounding factor $z$ for the source and target distributions.
% LA is tuned for worst group
% TTLSA is tuned for average gropu
We define $m_s=\max(\pi_s)$ and $m_t=\max(\pi_t)$
as the maximum probability of a $(y, z)$ group in the source and
target distributions.
The difference between these values reflects the degree of distribution shift.
%(LA is "logit adjustment".)
}
\label{tab:facebook}
\end{table}

\eat{
\begin{table}[h]
\centering
% \tiny
\begin{adjustbox}{width=\textwidth}
\begin{tabular}{l|c|c|c|c|c||c}
   Data & $m_s$ & $m_t$ & ERM  & gDRO & SUBG   & LA/TTLSA  \\
\hline
CelebA     &  0.44 & 0.49
& 80.83  (1.46) / {\bf 95.93} (0.03)
& {\bf 87.36} (0.47) / 94.68 (0.07)
& {\bf 87.10} (1.26) / 93.44 (0.19)
& 84.72 (0.58) / 95.52 (0.08)
\\
Waterbirds  & 0.73 & 0.39
& 85.78 (0.24) / 93.19 (0.16)
& 87.98 (0.86) / 93.06 (0.62)
& {\bf 88.87} (0.14) / 93.48 (0.11)
& {\bf 88.38} (0.36)  / 94.64 (0.40)
\\
MultiNLI  &  0.49 & 0.49
& 68.60 (0.40) / {\bf 82.70} (0.02)
& {\bf 76.79} (1.24) / 81.16 (0.07)
& 67.89 (0.91) / 72.15 (0.25)
& {\bf 76.33} (1.45) / 82.00 (0.50)
\\
CC  & 0.55 & 0.65
& 68.16 (1.03) / {\bf 88.00} (0.03)
& {\bf 79.66} (0.17) / 84.46 (0.43)
& 76.56 (0.25) / 79.56 (0.77)
& {\bf 79.27} (1.17) / 78.45 (1.10)
\\
\end{tabular}
\end{adjustbox}
\caption{Simplified version.
Accuracy of the worst / average $(y,z)$ group 
on the benchmark datasets.
%$\rho_s$ and $\rho_t$ are the empirical correlation coefficients between the label $y$ and the confounding factor $z$ for the source and target distributions.
The LA/TTLSA method is tuned for 
{\bf worst} group performance.
We define $m_s=\max(\pi_s)$ and $m_t=\max(\pi_t)$
as the maximum probability for the source and
target distributions.
The difference between these values reflects the nature and degree of distribution shift.
%(LA is "logit adjustment".)
}
\label{tab:facebook3}
\end{table}
}

Overall, the results show that TTLSA offers a unified approach to achieve a variety of robustness objectives.
In terms of average group accuracy, 
we find that the  performance of TTLSA 
relative to ERM
 depends on the nature of the shift.
%We find that the majority  $(y,z)$ meta-label
%does not change between source and target,
%although it may get upweighted or downweighted,
%depending on the dataset.
For CelebA and MultiNLI,  there is no significant distribution shift, so
TTLSA is similar to ERM, as expected.
For Waterbirds, 
the majority class in the source becomes much less common in the target
(reflecting the fact that the rare combinations of
water-birds on land  and land-birds on water
become more frequent).
TTLSA is able to adapt to this, and outperforms ERM.
For CivilComments, the majority class becomes even more common in the target distribution.
Although TTLSA can adapt to this change,
it cannot match the fact that ERM has been 
 ``rewarded'' for learning a representation that 
 is optimized for a single majority class.
 %relies on $z$ for predicting $y$.
 In terms of worst-group accuracy, the unadapted LA baseline (which is a special case of TTLSA where we do not use EM adaptation) achieves competitive performance to the dedicated gDRO and SUBG methods.
\section{Conclusions and future work}
\label{sec:discussion}

We have shown that adapting to changes in the  nuisance factors
$Z$ can give better results than using classifiers
that are designed to be invariant to such changes.
However, 
a central weakness of our approach is that it requires that the generative distribution $p(\vx | y, z)$ be preserved across domains.
This assumption becomes more plausible the more factors of variation are included in $Z$.
However, as $Z$ becomes high dimensional, each step of our TTLSA method, as well as access to appropriate data, becomes more challenging.
In this vein, another weakness is that we require access to labeled examples of $Z$ during training.
In the future, we would like to relax this assumption,
potentially by using  semi-supervised
methods \citep[c.f.,][]{Sohoni2021,Lokhande2022,Nam2022ssa}
that combine small fully labeled datasets
with large partially labeled datasets.\footnote{
We have conducted a preliminary experiment in which
 we fit $\psource(z|\vx)$ on a small fully
labeled subset, $\data_s^{xyz}$,
and then use this to
impute the missing $z$ values
on the larger $\data_s^{xy}$.
We then train a new model 
$\psource(y,z|\vx)$
on the soft predicted
$z$ labels and the hard observed $y$ labels.
We get good results on the worst group benchmarks
even when up to 90\% of the $z$ labels are missing (see \cref{sec:partialGroupLabel} in the supplementary material).
However, we leave more detailed evaluation of this method  to future work.
}
We also plan to explore the use of fully unsupervised estimates of the
confounding factors $Z$, based on generative models,
or by leveraging multiple source domains, 
similar to \citep{Jiang2022}.

\eat{
One final weakness is that we rely on the source domain classifier $\psource(y,z|\vx)$ to be well calibrated, both to evaluate the reweighting in \eqref{eq:posterior}, and so we can apply the EM procedure to $\ptarget(y,z)$.
Interestingly, we find that the latter operation is more sensitive to calibration, so alternative approach method of moments estimators proposed in 
\citet{Lipton2018} or \citet{Garg2020shift},
could potentially be more robust (albeit less statistically efficient than our MLE method).
%\todoqingyao{\citet{Alexandari2020} argues that EM/MLE is better than BBSL and RLLS, both of which are moment-matching methods}
}

We discuss potential societal impacts in the appendix.

%Another weakness of our approach is that we need
%access to labeled examples of $Z$ during training.
%In the future, we would like to relax this assumption,
%by using  semi-supervised
%methods (c.f, \citep{Sohoni2021,Lokhande2022,Nam2022ssa})
%that combine small fully labeled datasets
%with large partially labeled datasets.
%We also plan to explore the use of fully unsupervised estimates of the
%confounding factors $Z$, based on generative models,
%or by leveragig multiple source domains, 
%similar to \citep{Jiang2022}.
%

\section*{Acknowledgements}

We would like to thank
Andrew Sellergren at Google Health
for his help with the CheXpert CXR embeddings, and Jonathan Caton at Google TPU Research Cloud for providing computational resources.

\newpage
\appendix

\section{Derivation of the EM algorithm}
\label{sec:em-deriv}

In this section we describe how to 
estimate the label prior on the target distribution,
$\ptarget(y,z)=\ptarget(m)=\vpi_m$, using the unlabeled data $\data_t^{x}$.
There are several approaches to this,
including a moment matching method
called black box shift learning \citep{Lipton2018}
and an MLE approach based on
the EM algorithm \citep{Saerens2002}.
In \citep{Alexandari2020}, they show that the MLE approach is much better, provided the classifier is calibrated.
(See also \citep{Garg2020shift} for a unified analysis
of these two approaches.)

Since our augmented label space is expanded to include both class labels $y$ and meta-data $z$, the number of labels $M$ can be large, which can result in problems when computing the MLE. We therefore expand the previous approach to compute the MAP estimate,
using a Dirichlet prior of the form 
\be
\Dir(\vpi|\valpha)
 = \frac{1}{B(\valpha)}
 \prod_{m=1}^M \vpi_m^{\alpha_m - 1}
\ee
where $B(\valpha)$ is the normalization constant.
Note that the MLE solution can be recovered by setting 
$\valpha = \vone$, which represents a uniform prior.

The goal is to maximize the (unnormalized) log posterior of $\vpi$ given the unlabeled target data $\vX$:
\begin{align}
\loss(\vX;\vpi) &= \log \ptarget(\vpi, \vX) \\
&= \log \ptarget(\vX|\vpi) 
+ \log \Dir(\vpi|\valpha)  \\
&= \sum_{n=1}^N \log \ptarget(\vx_n|\vpi) + \log \Dir(\vpi|\valpha) \\
&= \sum_{n=1}^N \log \left[ \sum_{m=1}^M \vpi_m 
 \ptarget(\vx_n|m) \right] + \log \Dir(\vpi|\valpha)
\end{align}
The first term can be rewritten as
\begin{align}
\sum_n \log \left[ \sum_{m=1}^M \vpi_m  \psource(\vx_n|m) \right]
 &= \sum_n \log \left[ \sum_{m=1}^M \vpi_m  
 \frac{\psource(m|\vx_n) \psource(\vx_n)}{\psource(m)} \right] \\
 &= \sum_n \log \sum_m \frac{\psource(m|\vx)}{\psource(m)} \vpi_m + \const
\end{align}
This objective is a sum of logs of a linear function of $\vpi$,
as is the log prior. This needs to be maximized subject to the affine
constraints $\vpi_m \geq 0$ and $\sum_{m=1}^M \vpi_m=1$,
 so the problem is concave,
with a unique global optimum \citep{Alexandari2020}.

One way to compute this optimum is to use EM.
Let $\vpi^j$ be the estimate of $\vpi$ at iteration $j$;
we initialize with $\vpi_m^0 = \psource(m)$.
First note that
\begin{align}
\ptarget(\vx_n, m_n) 
&= \psource(\vx_n|m_n) \ptarget(m_n)
= \prod_{m=1}^M \left[ \psource(\vx_n|m) \vpi(m) \right]^{\ind{m_n=m}}
\end{align}
Hence the complete data log posterior is given by
\begin{align}
\loss(\vX, \vM;\vpi) &= \sum_{n=1}^N 
\sum_{m=1}^M \ind{m_n=m} \log [\vpi_m \psource(\vx_n|m)]
+ \log \Dir(\vpi|\valpha) 
\end{align}
so  the expected complete data log posterior is
\begin{align}
Q\left(\vpi, \vpi^{(j)}\right) &= E_{\vM} [\loss(\vX, \vM; \vpi) | \vX, \vpi^{(j)}] \\
    &= \sum_{n=1}^N 
\sum_{m=1}^M p(m_n=m | \vX, \vpi^j)  \log (\vpi_m \psource(\vx_n|m))
+ \log \Dir(\vpi|\valpha)  \\
  &= 
\sum_{m=1}^M N_m^j  \log (\vpi_m \psource(\vx_n|m))
+ \sum_{m=1}^M (\alpha_m-1) \log \vpi_m - \log B(\valpha) \\
&= 
\sum_{m=1}^M N_m^j \log \vpi_m
+ 
\underbrace{\sum_{m=1}^M N_m^j \log \psource(\vx_n|m)}_{\const}
+ \sum_{m=1}^M (\alpha_m-1) \log \vpi_m  + \const
\label{eqn:EMobj}
\end{align}
where we drop constants wrt $\vpi$,
and where we defined the expected counts to be
\begin{align}
N_m^j = \sum_{n=1}^N p(m_n=m|\vx_n,\vpi^j) 
\label{eqn:ecounts}
\end{align}
Hence in the E step we just need to compute
 the posterior responsibilities for each label:
\be
p(m_n=m|\vx_n,\vpi^j) 
= \frac{\vpi^{j}(m) \psource(\vx_n|m)}
{\sum_{m'=1}^M \vpi^{j}(m') \psource(\vx_n|m')}
= \frac{\vpi^{j}(m) \psource(m|\vx_n)/\psource(m)}
{\sum_{m'=1}^M \vpi^{j}(m') \psource(m'|\vx_n)/\psource(m')}
\label{eqn:Estep}
\ee
We plug this into \cref{eqn:ecounts}
and then maximize \cref{eqn:EMobj},
using a Lagrange multiplier to enforce the sum to one constraint.
We then get the following
(see e.g., Sec 4.2.4 of \cite{book1} for the derivation):
\be
\hat{\vpi}_m^{j+1} = \frac{\tilde{N}_m^j}
{\sum_{m'=1}^M \tilde{N}_{m'}^j}
\ee
where $\tilde{N}_m^j$ are the prior pseudo counts 
plus the expected empirical counts:
\begin{align}
\tilde{N}_m^j = N_m^j + \alpha_m-1
\end{align}

At convergence, we have
\be
\ptarget(y,z) = \hat{\vpi}_{y,z}^J
\ee
If we assume that the class label prior is constant,
and only the distribution of auxiliary labels  has changed,
then we can write
\be
\ptarget(y,z) = \psource(y) \ptarget(z|y)
\ee
where
\be
\ptarget(z|y) = \frac{\ptarget(y,z)}{
\sum_{z'} \ptarget(y,z')}
\ee
However, we do not make this fixed label assumption in our experiments.

\eat{
\subsection{Modeling issues}

It remains to specify the form of $\psource(y,z|\vx)$.
A simple choice is to use
\be
\psource(y,z|\vx) = \psource(z|\vx) \psource(y|z,\vx)
\ee
This is similar to a mixture of experts model,
where $\psource(z|\vx)$ is the weight given to expert $z$,
and $\psource(y|z,\vx)$ is the corresponding expert.
Thus we see that the $z$ feature "modulates" the
$\vx \ra y$ mapping.
This can be implemented by some kind of neural network with two output "heads", but where the $z$ head is fed back into the $y$ head.

}

\clearpage
\section{Datasets}
\label{supp:data}

In this section we discuss the datasets in more detail.

\subsection{\cmnist}

\begin{figure}
\centering
\begin{minipage}{.24\linewidth}
\includegraphics[width=\textwidth]{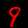}
\end{minipage}
\begin{minipage}{.24\linewidth}
\includegraphics[width=\textwidth]{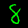}
\end{minipage}
\caption{
Samples from ColoredMNIST.
(a): $y=1$, $z=0$. 
(b) $y=1$, $z=1$.
}
\label{fig:mnist-sample}
\end{figure}

We show some sample images in \cref{fig:mnist-sample}.

\subsection{\chexpert}

\begin{figure}
\centering
\begin{minipage}{.34\linewidth}
\includegraphics[width=\textwidth]{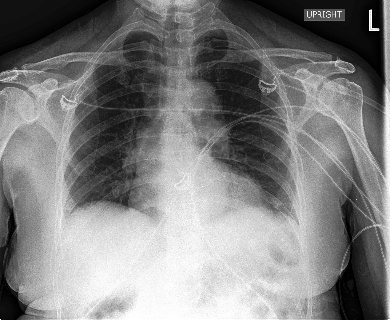}
\end{minipage}
\begin{minipage}{.34\linewidth}
\includegraphics[width=\textwidth]{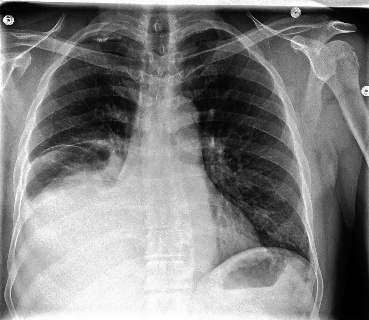}
\end{minipage}
\caption{Samples from CheXpert. \textbf{Left}: Female patient without effusion. 
\textbf{Right}: Male patient with effusion.
}
\label{fig:chexpert-sample}
\end{figure}

We show some sample images in \cref{fig:chexpert-sample}.
We list all the target attributes in
\cref{tab:chexpert-baseline}.
To test the difficult of each task,
we train a logistic regression model
for each attribute on the embeddins.
(We get similar results using an MLP.)
 The resulting AUC scores are shown in  \cref{tab:chexpert-baseline}.
This shows we can reliably predict all the attributes  from the embeddings.
The table also shows the marginal distribution of each attribute.
Many labels are highly skewed, which means accuracy would be a poor measure of the predictive performance.

Interestingly, we see that we can predict sex with an AUC of 0.973,
which is higher than the AUC for effusion (0.861).
To understand why, note that we only use frontal scans;
consequently breasts are often visible in female patients,
and this is often easier to detect visually than detecting
the disease itself (see \cref{fig:chexpert-sample}),
providing a possible ``shortcut'' for models to exploit.

\begin{longtable}[]{@{}lll@{}}
\toprule
\textbf{Attribute} & \textbf{AUC} & \textbf{Prob.} \\
\midrule
\endhead
    NO\_FINDING & 0.873 & 0.909 \\
    ENLARGED\_CARDIOMEDIASTINUM & 0.652 & 0.942 \\
    CARDIOMEGALY & 0.843 & 0.867 \\
    AIRSPACE\_OPACITY & 0.711 & 0.480 \\
    LUNG\_LESION & 0.761 & 0.963 \\
    PULMONARY\_EDEMA & 0.848 & 0.696 \\
    CONSOLIDATION & 0.683 & 0.911 \\
    PNEUMONIA & 0.742 & 0.973 \\
    ATELECTASIS & 0.694 & 0.815 \\
    PNEUMOTHORAX & 0.883 & 0.875 \\
    EFFUSION & 0.861 & 0.508 \\
    PLEURAL\_OTHER & 0.752 & 0.987 \\
    FRACTURE & 0.784 & 0.962 \\ 
    SUPPORT\_DEVICES & 0.900 & 0.420 \\
    GENDER & 0.973 & 0.586 \\
    AGE\_AT\_CXR & 0.914 & 0.492 \\
    PRIMARY\_RACE & 0.731 & 0.459 \\
    ETHNICITY & 0.681 & 0.728 \\
\bottomrule
\caption{Metrics for all the attributes in the CheXpert dataset.
(a) AUC using Logistic Regression on CXR embeddings.
(b) Baseline prior probability for each attribute,
illustrating the severe class imbalance for many attributes.
}
\label{tab:chexpert-baseline}
\end{longtable}

\clearpage
\section{Extra results}
\label{supp:results}

In this section, we include some extra experimental results.

\subsection{\cmnist using gradient boosted tree classifier}
\label{sec:treeClassifier}

\begin{figure*}
    \centering
    \includegraphics[width=\textwidth]{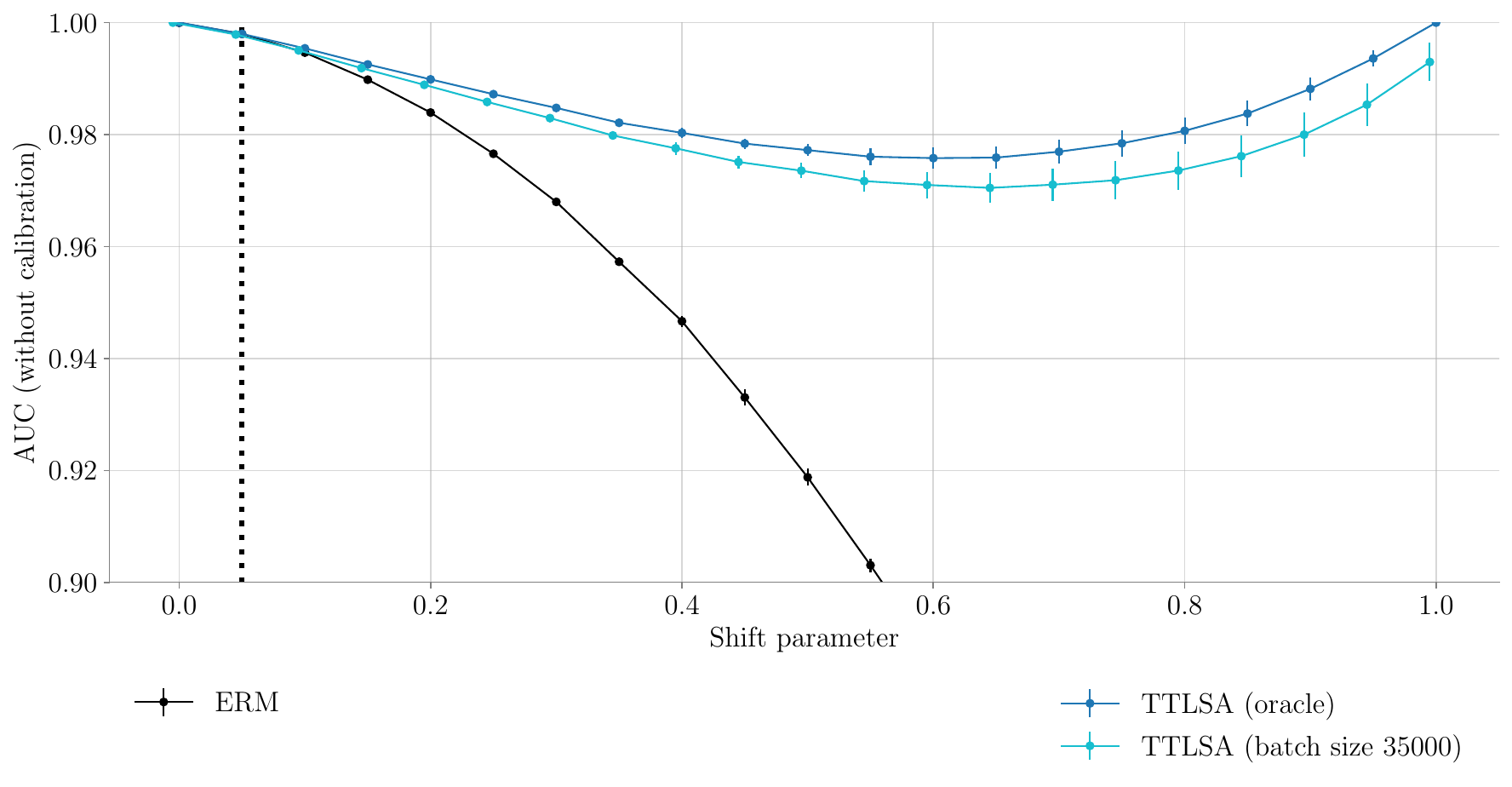}
    \caption{Performance on \cmnist using an uncalibrated tree classifier. TTSLA still improves the performance of the base model.}
    \label{fig:mnist-noise0-domain1-tree}
\end{figure*}

In \cref{fig:mnist-noise0-domain1-tree},
we show the results of various methods on the \cmnist dataset, where we use a 
Gradient Boosting Classification Tree
as our base classifier, instead of a DNN.
 In particular, we use the {\tt HistGradientBoostingClassifier} from scikit-learn~\citep{scikit-learn} with default parameters.
The results are qualitatively similar to the DNN case.

\subsection{The benefits of calibration}
\label{supp:calibration}

\begin{figure*}
\centering
\includegraphics[width=\textwidth]{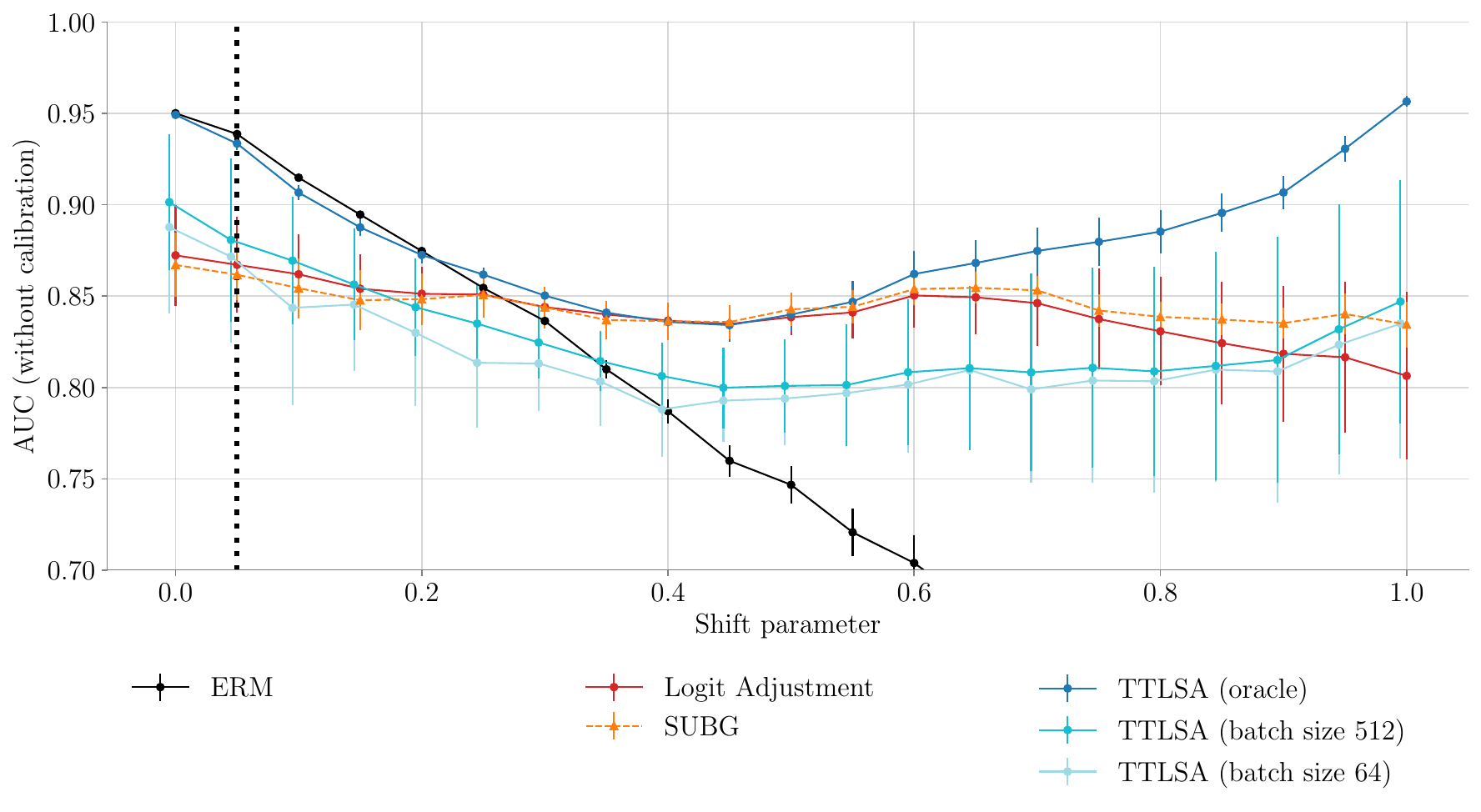}
\caption{
Performance across target domains on \chexpert embeddings, following the setup of Figure~\ref{fig:mnist-noise0-domain1-cali}.
(a) Results using calibration. Performance  mirrors those in Figure~\ref{fig:mnist-noise0-domain1-cali}.
(b) Results without calibration.
We see that calibration both improves performance and decreases variability between runs.
}
\label{fig:chexpert-embedding-domain1-nocali}
\end{figure*}

In \cref{fig:chexpert-embedding-domain1-nocali} we show the results on \chexpert if we remove the calibration step for our base classifier.
Compared to \cref{fig:chexpert-embedding-domain1-cali},
we see that the overall AUC of all the methods is worse, and the variance is larger.
\footnote{
For a binary classification problem, calibration will not change the AUC, but since we derive the posterior over class labels by marginalizing a 4-way joint,
$p(y|\vx) = \sum_{z=0}^1 p(y,z|\vx)$,
calibration can help.
}
However, the rank ordering of the methods is the same.
It is notable that a large gap opens up between the Oracle curve and the TTLSA implementations.
This suggests that calibration primarily improves estimation of $\ptarget(y,z)$ estimation via EM, because the Oracle curve in this subfigure corresponds to using the correct weights with the uncalibrated $\psource(y,z|\vx)$ model.

\eat{
\begin{figure*}
    \centering
    \includegraphics[width=\textwidth]{\figdir/mnist-domain1-noise0-cali1000_auc_minor.pdf}
    \caption{Performance on \cmnist (batch size ablation).}
    \label{fig:mnist-noise0-domain1-cali-ablation}
\end{figure*}

\begin{figure*}
    \centering
    \includegraphics[width=\textwidth]{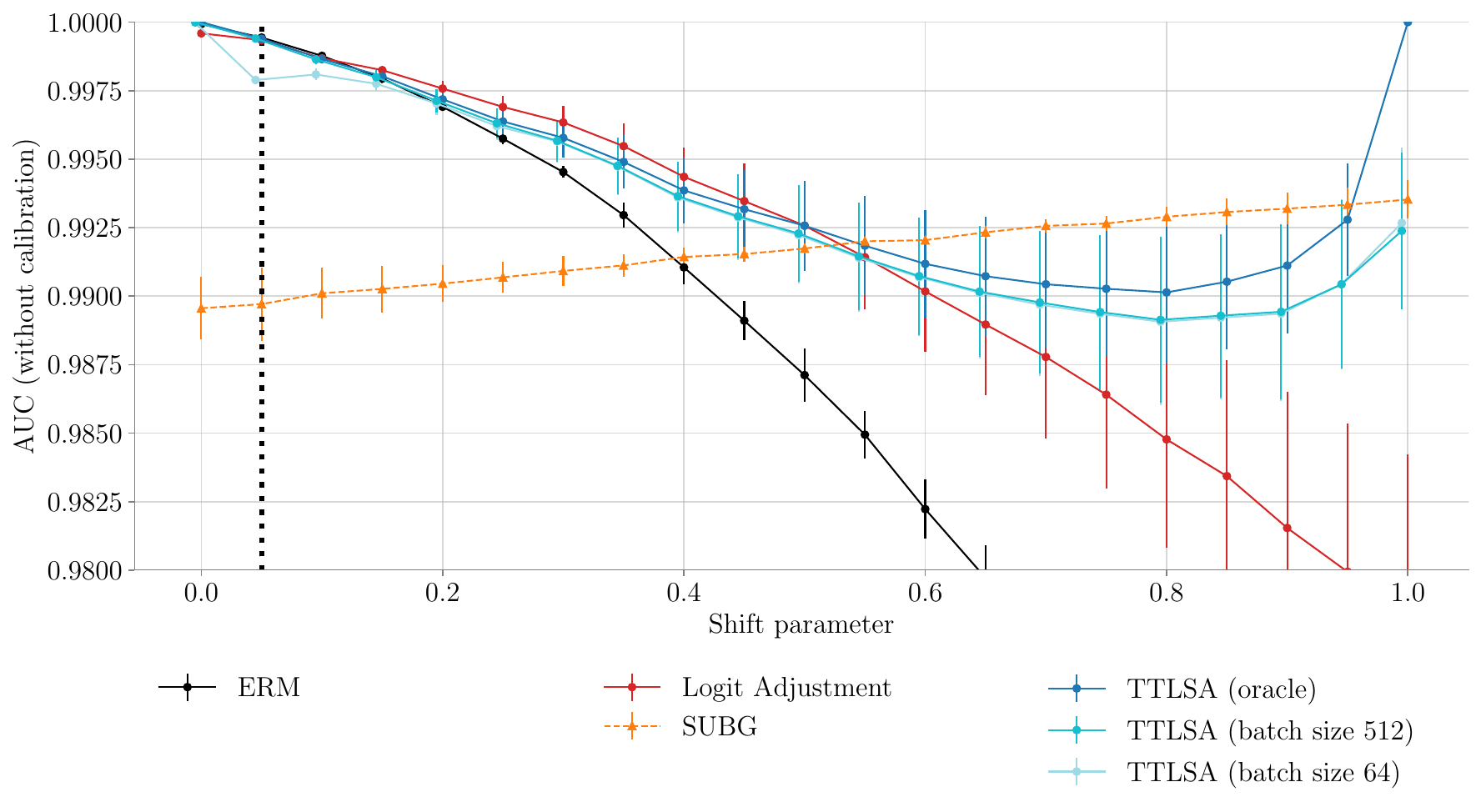}
    \caption{Performance on \cmnist without calibration MNIST.}
    \label{fig:mnist-noise0-domain1-nocali}
\end{figure*}

\begin{figure*}
    \centering
    \includegraphics[width=\textwidth]{\figdir/mnist-domain1-noise0-cali0_auc_minor.pdf}
    \caption{Performance on \cmnist without calibration (batch size ablation).}
    \label{fig:mnist-noise0-domain1-nocali-ablation}
\end{figure*}
}

\subsection{\chexpert using CNN on raw pixels}
\label{sec:chexpertPixels}

\begin{figure*}
\centering
\includegraphics[width=0.8\textwidth]{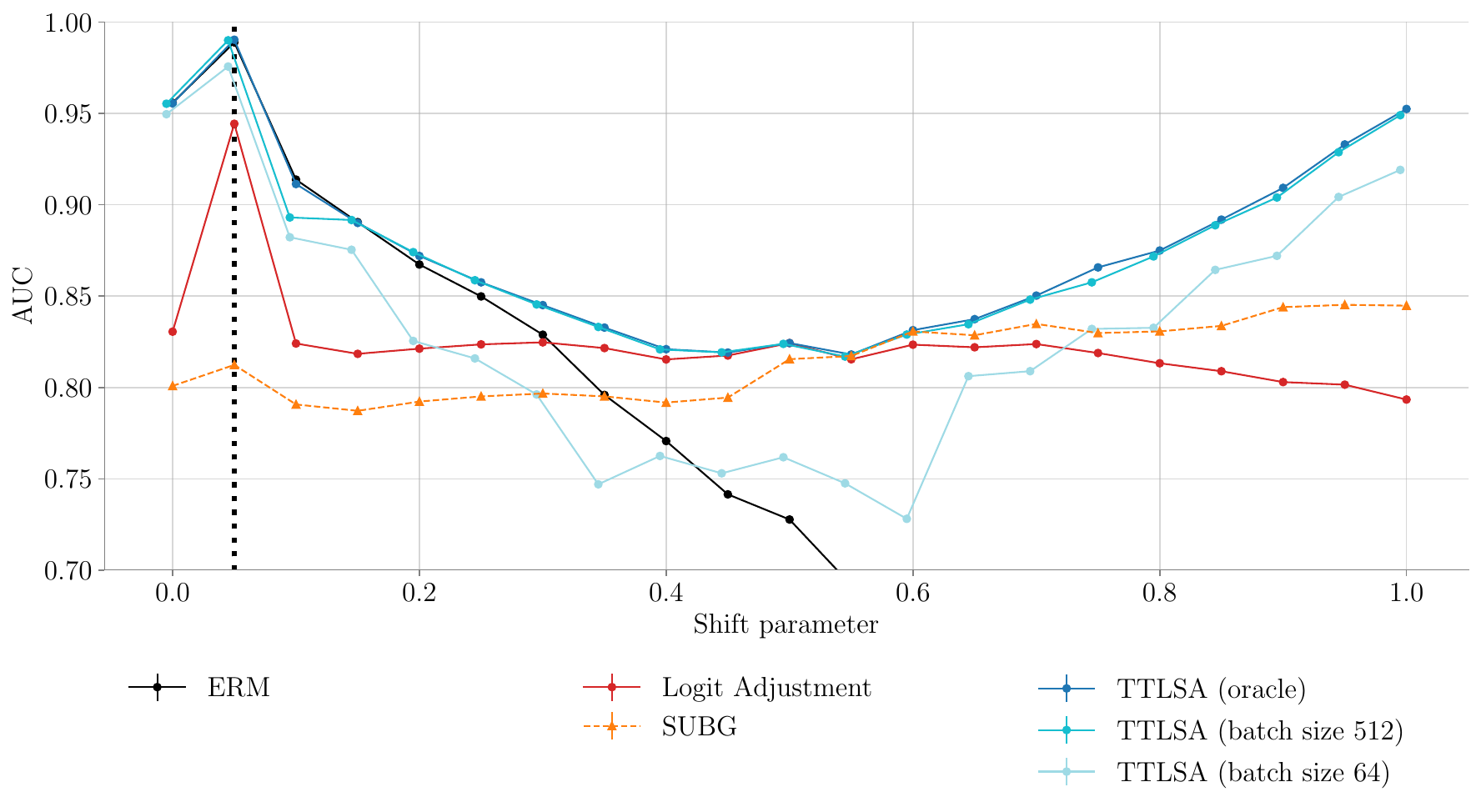}
\caption{
Performance on \chexpert using raw image (pixel) input instead of embeddings. These results are with calibration.
}
\label{fig:chexpert-pixel-domain1}
\end{figure*}

In \cref{fig:chexpert-pixel-domain1} we show the result of various methods when applied to \chexpert images, as opposed to using embeddings. 
We  use  a ResNet-50 that was pretrained on Imagenet, which we then fine tune on \chexpert images by replicating the gray-scale image along all 3 RGB channels.
The qualitative conclusions are the same as in the embedding case.

\subsection{More results on the benchmark datasets}
\label{supp:facebook}

In \cref{tab:facebook-per-class-min} and \cref{tab:facebook-per-class-avg}
we report the 
per-group accuracy on the benchmark datasets.

\begin{table}[h]
\centering
\begin{adjustbox}{width=\textwidth}
\begin{tabular}{l|c|c|c|c|c|c|c}
 &  & \multicolumn{6}{c}{Group label $(Y, Z)$} \\
Data & Method & (0, 0) & (0, 1) & (1, 0) & (1, 1) & (2, 0) & (2, 1) \\
\hline
CelebA & ERM & 86.71 (0.67) & 92.65 (0.79) & 96.80 (0.18) & 80.83 (1.46) \\
& gDRO & 92.71 (0.05) & 92.33 (0.09) & 92.63 (0.35) & 87.36 (0.47) \\
& SUBG & 91.76 (0.22) & 91.91 (0.55) & 90.96 (0.33) & 87.22 (1.38) \\
& LA & 91.43 (0.11) & 94.77 (0.08) & 95.88 (0.12) & 84.72 (0.58) \\
& TTLSA & 97.59 (0.18) & 98.78 (0.03) & 80.70 (1.22) & 51.25 (0.27) \\
Waterbirds & ERM & 98.95 (0.13) & 86.57 (0.48) & 86.02 (0.13) & 95.76 (0.16) \\
& gDRO & 93.46 (0.22) & 88.00 (0.88) & 90.15 (0.10) & 92.06 (0.34) \\
& SUBG & 90.76 (0.69) & 88.96 (0.19) & 91.28 (0.35) & 91.36 (0.24) \\
& LA & 94.43 (1.63) & 88.38 (0.36) & 91.32 (0.43) & 93.15 (0.54) \\
& TTLSA & 94.59 (0.44) & 93.68 (0.73) & 95.72 (0.29) & 97.12 (0.19) \\
MultiNLI  & ERM & 80.75 (0.79) & 94.94 (0.11) & 83.18 (0.47) & 78.05 (1.31) & 81.98 (0.48) & 68.60 (0.40) \\
& gDRO & 80.36 (0.63) & 85.27 (0.25) & 82.48 (0.59) & 81.21 (1.30) & 79.39 (1.34) & 76.87 (1.28) \\
& SUBG & 69.63 (0.17) & 82.85 (0.19) & 74.39 (0.21) & 79.68 (0.17) & 69.84 (0.48) & 68.40 (1.33) \\
& LA & 81.63 (1.15) & 87.79 (1.99) & 84.36 (0.89) & 80.95 (2.03) & 78.77 (1.09) & 76.33 (1.45) \\
& TTLSA & 80.24 (0.87) & 94.74 (0.58) & 81.73 (2.45) & 73.90 (1.72) & 82.40 (1.49) & 63.76 (2.15) \\
CivilComments & ERM & 92.23 (0.42) & 90.38 (0.46) & 68.57 (1.07) & 68.32 (0.97) \\
& gDRO & 83.94 (0.70) & 79.92 (0.33) & 80.97 (0.63) & 81.09 (0.42) \\
& SUBG & 79.79 (0.56) & 79.14 (0.34) & 82.52 (0.46) & 76.56 (0.25) \\
& LA & 84.45 (0.16) & 79.27 (1.17) & 83.00 (0.95) & 84.20 (0.99) \\
& TTLSA & 85.53 (1.35) & 74.94 (1.96) & 84.34 (2.36) & 84.61 (2.21) \\
\end{tabular}
\end{adjustbox}
\caption{Per-group accuracy on the benchmark datasets, where model selection is based on the worst $(Y, Z)$ group accuracy on a validation set.
Numbers in parentheses signify the standard error calculated based on 4 replication runs.
}
\label{tab:facebook-per-class-min}
\end{table}

\begin{table}[h]
\centering
\begin{adjustbox}{width=\textwidth}
\begin{tabular}{l|c|c|c|c|c|c|c}
 &  & \multicolumn{6}{c}{Group label $(Y, Z)$} \\
Data & Method & (0, 0) & (0, 1) & (1, 0) & (1, 1) & (2, 0) & (2, 1) \\
\hline
CelebA & ERM & 96.54 (0.21) & 99.58 (0.05) & 86.47 (0.89) & 40.28 (2.24) \\
& gDRO & 95.48 (0.14) & 96.76 (0.14) & 87.10 (0.34) & 68.75 (1.14) \\
& SUBG & 95.45 (0.39) & 95.93 (0.44) & 79.81 (1.30) & 67.64 (4.66) \\
& LA & 95.25 (0.55) & 98.96 (0.49) & 88.77 (1.81) & 43.47 (12.05) \\
& TTLSA & 97.68 (0.13) & 98.99 (0.07) & 80.21 (1.04) & 47.36 (1.81) \\
Waterbirds & ERM & 99.42 (0.11) & 90.27 (1.24) & 80.61 (2.51) & 94.16 (0.84) \\
& gDRO & 97.04 (1.44) & 92.84 (1.07) & 83.84 (2.47) & 89.10 (0.82) \\
& SUBG & 96.98 (0.29) & 95.88 (0.42) & 82.87 (1.51) & 83.33 (1.86) \\
& LA & 98.23 (0.15) & 92.42 (0.36) & 85.98 (1.06) & 92.91 (0.19) \\
& TTLSA & 99.06 (0.11) & 93.61 (1.04) & 87.66 (0.40) & 95.02 (0.92) \\
MultiNLI  & ERM & 82.43 (0.06) & 95.47 (0.08) & 83.62 (0.03) & 77.14 (0.16) & 80.45 (0.09) & 67.36 (0.54) \\
& gDRO & 80.37 (0.82) & 86.32 (0.64) & 81.06 (0.72) & 78.22 (0.60) & 81.22 (0.22) & 78.83 (0.29) \\
& SUBG & 68.30 (2.00) & 83.95 (2.28) & 75.72 (1.68) & 79.40 (1.37) & 69.91 (1.55) & 66.44 (2.23) \\
& LA & 82.74 (0.06) & 92.92 (0.34) & 83.97 (0.44) & 79.88 (0.55) & 79.77 (0.34) & 71.49 (0.95) \\
& TTLSA & 81.86 (0.19) & 96.52 (0.11) & 83.89 (0.37) & 76.07 (0.71) & 80.51 (0.17) & 56.60 (1.36) \\
CivilComments & ERM & 96.00 (0.38) & 95.63 (0.53) & 55.27 (1.88) & 52.21 (2.43) \\
& gDRO & 89.59 (0.68) & 86.60 (0.86) & 71.56 (1.64) & 71.94 (1.33) \\
& SUBG & 81.43 (1.09) & 80.67 (1.28) & 80.80 (1.34) & 76.05 (0.44) \\
& LA & 84.45 (0.16) & 79.27 (1.17) & 83.00 (0.95) & 84.20 (0.99) \\
& TTLSA & 91.53 (0.58) & 87.18 (1.70) & 70.14 (2.69) & 71.32 (2.82) \\
\end{tabular}
\end{adjustbox}
\caption{Per-group accuracy on the benchmark datasets, where model selection is based on the average $(Y, Z)$ group accuracy on a validation set.
Numbers in parentheses signify the standard error calculated based on 4 replication runs.
}
\label{tab:facebook-per-class-avg}
\end{table}

\eat{
\begin{table}[h]
\centering
% \tiny
\begin{adjustbox}{width=\textwidth}
\begin{tabular}{l|c|c|c|c|c||c|c}
   Data & $m_s$ & $m_t$ & ERM  & gDRO & SUBG & LA  & TTLSA  \\
\hline
CelebA     &  0.44 & 0.49
& 80.83  (1.46) / 95.93 (0.03)
& 87.36 (0.47) / 94.68 (0.07)
& 87.10 (1.26) / 93.44 (0.19)
& 84.72 (0.58) / 95.38 (0.09)
& 51.25 (0.27) / 95.55 (0.09)
\\
Waterbirds  & 0.73 & 0.39
& 85.78 (0.24) / 93.19 (0.16)
& 87.98 (0.86) / 93.06 (0.62)
& 88.87 (0.14) / 93.48 (0.11)
& 88.38 (0.36) / 94.02 (0.23)
& 93.65 (0.73) / 95.23 (0.34)
\\
MultiNLI  &  0.49 & 0.49
& 68.60 (0.40) / 82.70 (0.02)
& 76.79 (1.24) / 81.16 (0.07)
& 67.89 (0.91) / 72.15 (0.25)
& 76.33 (1.45) / 82.54 (0.05)
& 63.76 (2.15) / 82.60 (0.04)
\\
CC  & 0.55 & 0.65
& 68.16 (1.03) / 88.00 (0.03)
& 79.66 (0.17) / 84.46 (0.43)
& 76.56 (0.25) / 79.56 (0.77)
& 79.27 (1.17) / 80.99 (0.67)
& 74.94 (1.96) / 85.03 (0.71)
\\
\end{tabular}
\end{adjustbox}
\caption{Accuracy of the worst / average $(y,z)$ group 
on the benchmark datasets
using all methods separately.
%$\rho_s$ and $\rho_t$ are the empirical correlation coefficients between the label $y$ and the confounding factor $z$ for the source and target distributions.
We define $m_s=\max(\pi_s)$ and $m_t=\max(\pi_t)$
as the maximum probability for the source and
target distributions.
The difference between these values reflects the degree of distribution shift.
In all datasets, the majority $(y, z)$ group is the same across the source and target distributions.
%(LA is "logit adjustment".)
}
\label{tab:facebook-non-combined}
\end{table}
}

\subsection{Training with partial group labels}
\label{sec:partialGroupLabel}

In this section, we evaluate an extension of our method where not all training samples have group labels $z$.
In particular, we first train an ERM model to predict $z$ on samples with group labels $z$, calculate $p(z|x)$ for training samples with missing $z$,
and then fit a new $p(y,z|x)$ model on the 
augmented data. 
In particular, we represent each $(y,z)$ target as a one-hot vector when $z$ is known, and use a soft (predicted) encoding when $z$ is unknown. We train with cross entropy loss.
The use of soft labels  may have the benefits of self-distillation~\cite{pham2022revisiting}.
%The imputation model ignores $y$ labels.
The validation set is always fully labeled for the purpose of hyperparameter tuning.

The results (on the 4 benchmark datasets)
are shown in \cref{tab:facebook-per-imputation}.
The accuracy barely drops as missingness increases, which means our method is robust to the deficiency in group labels $z$.
%In practice, this trend is likely to hold true because $z$ labels must be easy to learn to become a ``shortcut''.

\begin{table}[h]
\centering
\begin{adjustbox}{width=\textwidth}
\begin{tabular}{l|c|c|c|c|c}
 & \multicolumn{5}{c}{Missingness} \\
Data & 0 & 0.5 & 0.75 & 0.875 & 0.9375 \\
\hline
CelebA & 84.72 / 95.55 & 78.33 / 95.68 & 77.78 / 95.37 & 79.44 / 94.44 & 77.22 / 95.20 \\
Waterbirds & 88.38 / 95.23 & 87.63 / 93.98 & 88.79 / 94.41 & 88.65 / 94.67 & 91.28 / 95.05 \\
MultiNLI & 76.33 / 82.60 & 74.87 / 79.55 & 74.72 / 82.49 & 76.05 / 82.61 & 78.75 / 81.72 \\
CivilComments & 79.27 / 85.03 & 76.26 / 85.87 & 73.87  / 83.55 & 73.41 / 84.57 & 66.64 / 80.36 \\
\end{tabular}
\end{adjustbox}
\caption{Accuracy of the worst / average $(y, z)$ group on the benchmark datasets with partial training $z$ labels, where model selection is based on average $z$ accuracy.
The \textit{Missingness} columns stand for the proportion of training set with missing labels, e.g. 0.75 means only 25\% of the training samples have $z$ labels.
%For the convenience of the reader, we have included the last column from \cref{tab:facebook} as the first column.
}
\label{tab:facebook-per-imputation}
\end{table}

\clearpage
\section{Potential negative societal impacts}

The proposed method in this work yields a model that can adapt to a new distribution and improves the performance at test time by exploiting spurious correlations to create a label shift correction technique that adapts to changes in the marginal distribution $p(y, z)$ using unlabeled samples from the target domain. 
In this way, there are potential societal benefits to our method, especially when $z$ corresponds to a socially salient attribute, such as a protected class.
However, use cases of this type require caution, especially given the limitations discussed in \cref{sec:discussion}.
Further, as we discuss in a footnote in the main text, our method does not address concerns about cases where making decisions on the basis of $z$ is discouraged or forbidden for \emph{a priori} reasons.
Given these limitations, there is a potential that the existence of adaptation methods of this type could be used to downplay the potential dangers of misusing sensitive information in machine learning systems.
Here, we hope researchers and practitioners will instead acknowledge that, while beneficial use cases of $z$ information exist, (1) there is a need to validate empirically that a particular use of $z$ information is actually socially beneficial, and (2) there are valid reasons why one might want to avoid using $z$ information altogether.
Further, there is a potential risk that if the measurement quality of the labels $y,z$ shift across distributions, such that they measure distinct concepts, or exhibit substantially different noise properties (i.e., become biased, or exhibit more outliers), our framework might absorb them during adaptation and eventually the outcomes of the system might be biased as well.

\clearpage
\section{Invariance Equivalences and Conditions}
In this section, we review connections that have been established between risk invariance, ERM on balanced data, ``separation'' between a predictor $f(X)$ and the spurious factor $Z$, and worst-$(y,z)$-group performance.
These results are useful for understanding why the application of logit adjustment at training time often yields a predictor that exhibits approximately invariant risk across the test sets that we study in our experiments.

\subsection{Key Concepts}
\paragraph{Risk invariance}
A predictor is risk-invariant with respect to a loss function $\ell$ and a family of test distributions $\mathcal Q$ iff it has the same risk $E_Q[\ell(f(X), Y)]$ for each $Q \in \mathcal Q$.
The results we discuss apply to test distribution families that preserve both the generative distribution \emph{and} the label distribution of the source distribution; that is, $\mathcal Q$ is the set of distributions such that $Q(Y)=P(Y)$ and $Q(X \mid Y, Z)=P(X \mid Y, Z)$ for each $Q \in \mathcal Q$.
This formulation allows $Q(Z \mid Y)$ can change.
This is is the family is considered in \citet{Makar2022}  and \citet{Makar2022fairness}, and is called a ``causally compatible'' family in \citet{Veitch2021}, or a correlation shift in \citet{yibreaking}.

\paragraph{Pure spuriousness}
The data generating process in Figure 1 is purely spurious if there exists some sufficient statistic $e(X)$ such that (1) $\indep{Y}{X} \mid {e(X)}$ and (2) $\indep{e(X)}{Z} \mid Y$.
In words, if we know $e(X)$, there is no further dependence between $Y$ and $X$, and further, $e(X)$ does not depend on the spurious factor $Z$ except through $Z$'s marginal dependence with $Y$.
This is consistent with a causal model where the influence of $Y$ on $X$ is totally mediated by $e(X)$, and $Z$ has no causal effect on $e(X)$.

\citet{Veitch2021} coined the term ``purely spurious'' in a context of a full counterfactual model of data generation, to refer to data generating processes where the portions of $X$ that are causally related to $Y$ and $Z$ can be separated in a specific sense.
\citet{Makar2022} consider the special case of pure spuriousness in the context of the anti-causal model in Figure 1.
(They do not use the term ``purely spurious'' as the work in \citet{Veitch2021} was concurrent; \citet{Makar2022fairness} makes the connection explicit.)
Here, we use formalism from \citet{Makar2022} to present the idea to minimize conceptual overhead.

Note that when the data $X$ is rich, such as images are long passes of text, pure spuriousness is more plausible (or a better approximation to reality) because there is less possibility of descructive interfecence between $Y$ and $Z$ in the generation of $X$.
Specifically, the simplest examples where pure spuriousness fails are ones where $X$ is very low-content: e.g., $Y$ and $Z$ are binary, and $X := Y \texttt{ OR } Z$.

\paragraph{Separation}
Separation is a concept popularized in the literature on ML fairness \citep[][Chapter 3]{barocas-hardt-narayanan}, which stipulates that the predictor $f(X)$ should satisfy the the conditional independence $f(X) \indep Z \mid Y$.
When $Z$ is a sensitive attribute, this condition stipulates that the predictor $f(X)$ should contain no more information about $Z$ than one could glean from knowing $Y$ alone.

\paragraph{Data balancing}
\citet{idrissi2022simple} study predictors trained on data subsampled so that the $(Y, Z)$ distribution is uniform; they call this data-balancing.
\citet{Makar2022} and \citet{Makar2022fairness} study a similar predictors optimized on a similar ``ideal'' distribution, where $Q(Y, Z) = P(Y)P(Z)$ for some source distribution $P$.
This distribution does not ``balance'' the marginals of $Y$ and $Z$, but it eliminates the marginal correlation between $Y$ and $Z$.

\paragraph{Worst group performance}
\citet{Sagawa2020} define groups in terms of $(z, y)$ values.
The group conditional risk is $R_{z, y} = E_Q[\ell(f(X), Y) \mid Z=z, Y=y]$.
Note that for all families of test sets that we consider, the group-conditional risks are equal for all Q.
Worst group risk minimization attempts to minimize the group conditional risk of the worst subgroup.
\citet{Saerens2002} propose a distributionally robust optimization algorithm for performing this minimization.

\subsection{Connections}

In the purely spurious setting, there are several connections and near-equivalences between risk invariance, separation, optimality on balanced data, and worst group risk minimization.

\citet{yibreaking} establish that for label distribution preserving target families, a predictor $f(X)$ that satisfies separation $\indep{f(X)}{Z} \mid Y$ will have invariant risk across the family $\mathcal Q$ defined above. Notably, this result does \emph{not} require pure spuriousness.

Under pure spuriousness, the separation condition achieves a certain optimality.
\citet{Veitch2021}, Theorem 4.3 establishes that in the purely spurious case, the minimax optimal across the family $\mathcal Q$ satisfies separation $\indep{f(X)}{Z} \mid Y$.
Similarly, under pure spuriousness, \citet{Makar2022fairness}, Proposition 2, establishes that the optimal risk-invariant predictor satisfies separation.

Interestingly, this result establishes a connection between optimality under balanced data, separation, and optimal risk invariance.
Specifically, \citet{Makar2022}, Proposition 1 establishes that the optimal model for the ``ideal'' uncorrelated distribution for which $Q(Y, Z) = P(Y)P(Z)$ achieves risk invariance across the family $\mathcal Q$.
Thus, minimizing risk under a separation constraint targets a similar predictor to the predictor that one would target simply optimizing on balanced data.
\citet{Makar2022fairness} shows that the near-equivalence holds up empirically, such that learning algorithms targeted at efficiently learning the optimal predictor on balanced data can satisfy both risk invariance and separation criteria.

\citet{idrissi2022simple} establish that, empirically, models trained to minimize risk on balanced data also yield favorable worst-group performance, showing that subsamping can be particularly effective.
\citet{Sagawa2020} explore similar ideas, focusing on reweighting strategies, which both they and \citet{idrissi2022simple} find to work relatively poorly with neural models in the data regimes they study.
\citet{Sagawa2020} further establish that under certian convexity conditions, there does exist a reweighting of the data that optimizes worst-group performance, but provide a counterexample showing that this is not always the case with non-convex losses.

Based on the above results, in the purely spurious case, one can establish the following, for $\mathcal Q$ with a uniform distribution on $Y$:
\begin{enumerate}
\item There exists a predictor $f^*(X)$ that is optimal on the ideal balanced data, is the optimal risk-invariant predictor, and satisfies separation $\indep{f(X)}{Z} \mid Y$.
\item For all $Q \in \mathcal Q$, the group-specific risks are equal within labels, i.e., $E_Q[\ell(f^*(X), Y) \mid Y=y, Z=z] = E_Q[\ell(f^*(X), Y) \mid Y=y, Z=z']$ for all $y$.
\end{enumerate}
The latter fact does not imply that $f^*(X)$ also optimizes worst-group risk, but it does imply that the worst group cannot be the worst due to a spurious correlation between $Y$ and $Z$.
This is because, for a fixed label value $y$, the risks of $(y, z)$ subgroups are the same.

\medskip
\bibliographystyle{abbrvnat}
\bibliography{bib}

%%%%%%%%%%%%%%%%%%%%%%%%%%%%%%%%%%%%%%%%%%%%%%%%%%%%%%%%%%%%
\end{document}